\documentclass[letterpaper]{article} 

\usepackage{arxiv}  
\usepackage{times}  
\usepackage{helvet}  
\usepackage{courier}  
\usepackage{graphicx} 
\usepackage{color} 
\usepackage[dvipsnames,svgnames,table]{xcolor}
\usepackage[colorlinks]{hyperref}
\hypersetup{
    colorlinks=true,
    citecolor=brown,
    linkcolor=Mahogany,
    filecolor=orange,      
    urlcolor=orange,
    }

\usepackage{natbib}  
\usepackage{subcaption}
\usepackage{amsthm,amsmath,amssymb,,bm}
\usepackage{colortbl}
\usepackage{booktabs}
\usepackage{caption} 
\usepackage{algorithm}
\usepackage{algorithmic}
\usepackage{mdframed}
\usepackage{multirow} 
\usepackage{threeparttable}

\DeclareMathOperator*{\argmin}{arg\,min}

\newtheorem{theorem}{Theorem} 
\newtheorem{lemma}{Lemma} 
\newtheorem{prop}{Proposition}
\newtheorem{remark}{Remark}

\usepackage{newfloat}
\usepackage{listings}
\DeclareCaptionStyle{ruled}{labelfont=normalfont,labelsep=colon,strut=off} 
\floatstyle{ruled}
\newfloat{listing}{tb}{lst}{}
\floatname{listing}{Listing}
\title{Multimodal Variational Autoencoder: a Barycentric View}
\author{
    Peijie Qiu\textsuperscript{\rm 1}, Wenhui Zhu\textsuperscript{\rm 2}, Sayantan Kumar\textsuperscript{\rm 1}, Xiwen Chen\textsuperscript{\rm 3}, Jin Yang\textsuperscript{\rm 1}, Xiaotong Sun\textsuperscript{\rm 4}, Abolfazl Razi\textsuperscript{\rm 3}, Yalin Wang\textsuperscript{\rm 2}, Aristeidis Sotiras\textsuperscript{\rm 1\thanks{Corresponding author}}
}
\affiliations{
    \textsuperscript{\rm 1} Washington University in St. Louis, 
    \textsuperscript{\rm 2} Arizona State University, 
    \textsuperscript{\rm 3} Clemson University, 
    \textsuperscript{\rm 4} University of Arkansas \\
    \{peijie.qiu, sayantan.kumar, yang.jin, aristeidis.sotiras\}@wustl.edu, \\ \{wzhu59, ylwang\}@asu.edu, \{xiwenc, arazi\}@clemson.edu, xs018@uark.edu
}

\usepackage{bibentry}

\begin{document}

\maketitle

\begin{abstract}
Multiple signal modalities, such as vision and sounds, are naturally present in real-world phenomena. Recently, there has been growing interest in learning generative models, in particular variational autoencoder (VAE), to for multimodal representation learning especially in the case of missing modalities. The primary goal of these models is to learn a modality-invariant and modality-specific representation that characterizes information across multiple modalities. 
Previous attempts at multimodal VAEs approach this mainly through the lens of experts, aggregating unimodal inference distributions with a product of experts (PoE), a mixture of experts (MoE), or a combination of both. In this paper, we provide an alternative generic and theoretical formulation of multimodal VAE through the lens of barycenter. We first show that PoE and MoE are specific instances of barycenters, derived by minimizing the asymmetric weighted KL divergence to unimodal inference distributions. Our novel formulation extends these two barycenters to a more flexible choice by considering different types of divergences.
In particular, we explore the Wasserstein barycenter defined by the 2-Wasserstein distance, which better preserves the geometry of unimodal distributions by capturing both modality-specific and modality-invariant representations compared to KL divergence. Empirical studies on three multimodal benchmarks demonstrated the effectiveness of the proposed method.
\end{abstract}

\section{Introduction}
Multiple data types are naturally present together to characterize the same underlying phenomena in the real world. Multimodal representation learning is thus of interest across various fields, including computer vision, natural language processing, and the biomedical domain. However, understanding and interrelating different modalities is a challenging task due to the laboriousness of human annotations and the absence of certain modalities in practice.
These two factors pose a significant challenge to the application of unimodal and discriminative (supervised) representation learning methods to the multimodal case~\citep[see \textit{e.g.,}][]{karpathy2015deep,pham2019found,lin2023vision}. 

Therefore, we focus on the generative models for representation learning, which are typically considered as unsupervised, such as generative adversarial networks~\citep[GANs;][]{goodfellow2014generative} and variational autoencoders~\citep[VAEs;][]{vae}. In particular, we focus on VAEs for multimodal representation learning since VAEs are graphical probabilistic models capable of learning an explicit latent distribution, which has the potential to directly learn the joint distributions of multiple modalities~\cite{suzuki2016joint,baltruvsaitis2018multimodal}. Despite their nice probabilistic properties and the success in unimodal applications, the direct translation of VAEs to the multimodal case (\textit{e.g.,} feeding the multimodal data to VAEs) is challenging, as they struggle with handling missing modalities and performing cross-modal generations. Therefore, the design of multimodal VAEs seeks to form a modality-invariant and modality-specific latent representation by learning a joint latent distribution (so-called joint posterior) to aggregate the information from different modalities~\cite{ngiam2011multimodal,suzuki2016joint,baltruvsaitis2018multimodal}. The modality-specific and modality-invariant formulation naturally enables a cross-modal generation~\cite{shi2019variational}.
In addition, it can also handle missing modalities by directly sampling the learned joint posterior. 

The core objective of multimodal VAEs then revolves around how to approximate the joint posterior by aggregating the unimodal posterior, also known as unimodal inference distribution in VAEs. This typically involves finding a proper aggregation function.
However, such aggregation functions are challenging to identify due to the intractability of the true joint posterior. Previous explorations of multimodal VAEs addressed this challenge mainly through the lens of experts in statistics by aggregating unimodal inference distributions with a product of experts~\citep[PoE;][]{wu2018multimodal}, a mixture of experts~\citep[MoE;][]{shi2019variational}, or a combination of both~\citep[MoPoE;][]{sutter2021generalized}. Although empirical studies have shown their success for multimodal VAEs, theoretical analysis of their properties is still insufficient. 

In this paper, we provide a theoretical view of previous multimodal VAEs in a unified way through the lens of barycenter. The barycentric distribution is the mean distribution of a set of distributions, defined by minimizing the weighted sum of divergences to these distributions. Interestingly, we discovered that the distributions aggregated by PoE and MoE are barycenters by optimizing the reverse and forward Kullback-Leibler (KL) divergence, respectively. This directly provides an information-theoretic view of PoE and MoE, which reveals their intrinsic properties: PoE is zero-forcing (\textit{i.e.,} pushing the joint posterior biased towards certain modalities), while MoE is mass-covering (\textit{i.e.,} balancing all modalities).
However, the KL divergence does not define a metric space for probability measures, as it is asymmetric and unbounded. This motivates us to explore other divergence measures that are defined in metric space. In particular, we explored the Wasserstein barycenter~\cite{agueh2011barycenters} by optimizing the squared 2-Wasserstein distance, as it preserves the geometry of unimodal inference distributions in a geodesic space (whereas KL divergence focuses on pointwise differences). Leveraging the intricate geometry of the Wasserstein distance~\cite{wbgeometry}, the Wasserstein barycenter serves as the Fréchet means~\citep[see \textit{e.g.,}][]{frechetmean} within the space of probability measures. 

In summary, our contributions are threefold: 
 \textbf{i)} We introduce a novel and unified formulation for multimodal VAEs, where the aggregation of unimodal inference distributions is framed as solving the barycenter problem that minimizes certain divergence measures. This approach offers a theoretical framework to analyze intrinsic properties and enables a more flexible selection of aggregation functions for multimodal VAEs.
\textbf{ii)} We propose $\mathcal{WB}$-VAE, a novel multimodal VAE for representation learning that leverages the Wasserstein barycenter to aggregate unimodal inference distributions.
\textbf{iii)} Experiments on three benchmark datasets demonstrated the effectiveness of the proposed method compared to other state-of-the-art methods.

\section{Background and Related Work}

\subsection{Multimodal VAEs}
Prior multimodal VAEs can be roughly divided into two main categories: 
coordinated models and joint models. The former only learns the inference distributions from a single modality, while the latter learns the joint inference distributions across all modalities~\cite{baltruvsaitis2018multimodal,suzuki2022survey}. Accordingly, coordinated models~\cite{higgins2017scan,schonfeld2019generalized,korthals2019multi} strive to generate consistent inference results across all modalities. Although they can perform cross-modal generation, they may not effectively handle missing modalities as in joint models~\cite{wu2018multimodal,shi2019variational,sutter2020multimodal}. This is because they do not model the joint inference distribution of all modalities as in joint models. 

Here, we focus on joint models that can be applied to a wider spectrum of applications. Although there are some joint models that can handle missing modalities via a surrogate unimodal inference model~\cite{vedantam2017generative,korthals2019multi}, they typically face scalability issues. Hence, we consider joint models that can directly learn the joint inference distributions by aggregating unimodal inference distributions through an aggregation function.
Following this vein, \citet{wu2018multimodal} proposed an PoE-VAE (\textit{a.k.a.,} MVAE) by aggregating the unimodal distributions with a product of experts. Despite resulting in a sharper joint distribution, PoE-VAE is prone to focus on certain modalities while neglecting others. To mitigate this issue,~\citet{shi2019variational} proposed an MoE-VAE (\textit{a.k.a.,} MMVAE) by leveraging a mixture of experts. 
However, MoE-VAE does not produce a joint distribution that is sharper than any other expert: the precision of the joint inference distribution may not increase as the number of modalities increases. To take advantage of both PoE and MoE,~\citet{sutter2021generalized} proposed a generalized MoPoE-VAE, which first applies PoE and then MoE to all possible subsets of modalities. However, the previous attempts at joint models are limited to the perspective of experts in statistics. 

Although there are other multimodal VAEs~\cite{palumbo2023mmvae+,hirtlearning,yuan2024learning}, their focus is not on new aggregation functions. Instead, they are considered variants of PoE-VAE and MoE-VAE. In this paper, we provide a unified framework for aggregation functions from a barycentric view. In contrast to previous works that combined unimodal distribution aggregation with model parameter optimization~\cite{wu2018multimodal,shi2019variational,sutter2020multimodal,sutter2021generalized}, our barycentric formulation decouples these two steps. This enables a more flexible choice of barycenters for aggregating unimodal inference distributions (\textit{e.g.,} the Wasserstein barycenter, which we explore in this paper).

\subsection{Optimal Transport and Wasserstein distance} 
We briefly introduce optimal transport theory here to make this paper self-contained, since it will be used for the derivation of Wasserstein barycenter. 
Optimal transport (OT) seeks to find a transport map to move the mass from one distribution to another while minimizing the transport cost. Here, we consider Kantorovich's dual OT formulation~\cite{kantorovich1942translocation} instead of Monge's primal formulation~\cite{monge1781memoire}, as Monge's formulation is not symmetric.
For two probability measures\footnote{In a less rigorous sense, we use probability measures and probability distributions interchangeably, hereafter.} $P  \sim \mathcal{P}(\mathcal{X})$ and $Q  \sim \mathcal{P}(\mathcal{Y})$, with $\mathcal{P}(\mathcal{X})$ and $\mathcal{P}(\mathcal{Y})$ being the respective sets of probability distributions on them, Kantorovich's OT formulation is defined as

\begin{equation}
\inf\limits_{ \pi \in \prod(P, Q) } \int_{\mathcal{X} \times \mathcal{Y} } {c\left({x, y} \right)} d \pi(x, y) \nonumber,
\end{equation}
where $c: \mathcal{X} \times \mathcal{Y}$ is a cost function. The infimum is taken over the set of all transport plans $\pi \in \prod(P, Q) $, \textit{i.e.,} joint distributions on $\mathcal{X} \times \mathcal{Y}$ with marginals $P$ and $Q$. 

The $p$-Wasserstein distance is then the $p$-th root of the infimum of Kantorovich's OT formulation for a cost function $c(x,y) = |x-y|^p$:
\begin{equation}
    \mathcal{W}_p(P, Q) = \inf\limits_{\pi \in \prod(P, Q)} \left(\int_{\mathcal{X} \times \mathcal{Y} } |x-y|^p d \pi(x, y) \right)^{1/p}, \nonumber
\end{equation}
with $p=1$ being an earth mover's distance that is commonly used in many generative adversarial networks~\cite[see \textit{e.g.,}][]{arjovsky2017wassersteingan,gulrajani2017improved,miyato2018spectral}. In contrast, we focus on the 2-Wasserstein distance for deriving the Wasserstein barycenter in this paper, as its quadratic form allows for an analytic solution in the case of Gaussian distributions. For two Gaussian distributions $\mathcal{N}(\bm{\mu}_1, \bm{\Sigma}_1)$ and $\mathcal{N}(\bm{\mu}_2, \bm{\Sigma}_2)$, the squared 2-Wasserstein distance between them is solved analytically  ~\citep[see \textit{e.g.,}][]{knott1984optimal,givens1984class}: 
\begin{equation}
\begin{split}
     \mathcal{W}_2^2(\mathcal{N}(\bm{\mu}_1, &\bm{\Sigma}_1), \ \mathcal{N}(\bm{\mu}_2, \bm{\Sigma}_2)) = |\bm{\mu_1} - \bm{\mu_2}|_2^2 + \\
    &\text{Tr}(\bm{\Sigma}_1 + \bm{\Sigma}_2 - 2 (\bm{\Sigma}_1^{1/2} \bm{\Sigma}_2 \bm{\Sigma}_1^{1/2})^{1/2} ).
\end{split}
\end{equation}

\begin{figure}[!t]
    \centering
    \includegraphics[width=0.9\linewidth]{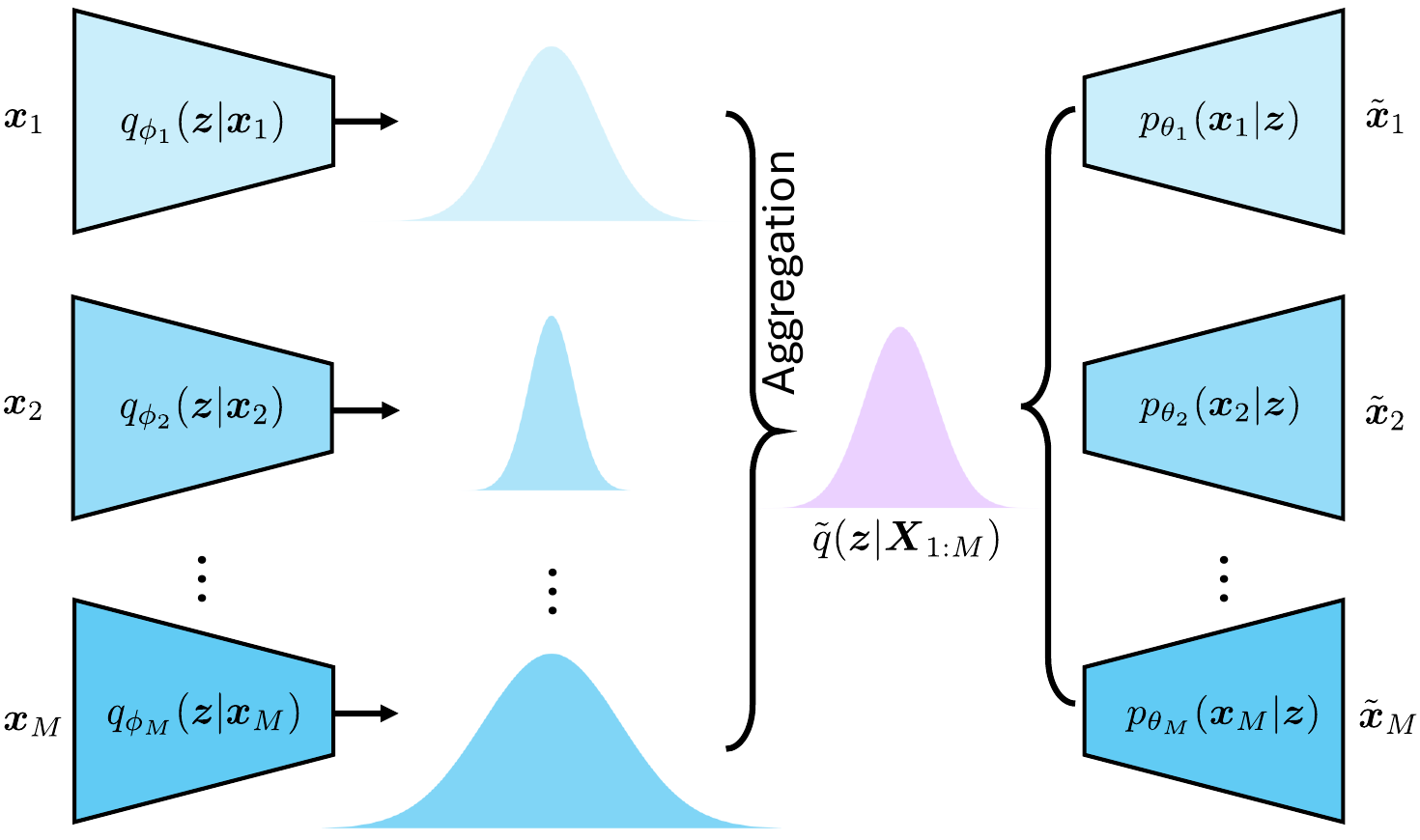}
    \caption{The overview of a multimodal VAE that takes $M$ modalities $\bm{X}_{1:M}=\{ \bm{x}_{j} \}_{j=1}^M$ as input and outputs the reconstructed input modalities $\Tilde{\bm{X}}_{1:M}=\{ \Tilde{\bm{x}}_{j}\}_{j=1}^M$. The multimodal VAE consists of $M$ probabilistic encoders $\{q_{\phi_j}(\bm{z}|\bm{x}_j)\}_{j=1}^{M}$ and decoders $\{p_{\theta_j}(\bm{x}_j|\bm{z})\}_{j=1}^{M}$.}
    \label{fig:graphical_abstract}
\end{figure}

\section{Method}

\subsection{Multimodal VAE: an Expert View}
Without loss of generality, we consider a dataset $\{\bm{X}^{(i)}_{1:M}\}_{i=1}^{N}$ containing $N$ number of independent and identically distributed (i.i.d.) samples, each of which consists of $M$ modalities: 
$\bm{X}^{(i)}_{1:M}=\{ \bm{x}_{1}^{(i)}, \cdots, \bm{x}_{M}^{(i)} \}$. Assuming the multimodal data can be generated by some random process involving a joint latent variable $\bm{z}$, the objective of a multimodal VAE is to maximize the log-likelihood of data over all $M$ modalities, given i.i.d. condition:
\begin{equation}\label{eqn:1}
\begin{split}
    \log p_{\theta}(\bm{X}^{(i)}_{1:M}) = D_{\text{KL}}(q_{\phi}(\bm{z} | \bm{X}^{(i)}_{1:M}) ||p_{\theta}(\bm{z}|\bm{X}^{(i)}_{1:M})) \\ + \mathcal{L}(\theta, \phi; \bm{X}^{(i)}_{1:M}),
\end{split}
\end{equation}
where $q_{\phi}(\bm{z} | \bm{X}^{(i)}_{1:M})$ is the approximate posterior parameterized by deep neural networks (\textit{i.e.,} the probabilistic encoders in VAEs), as the true posterior is intractable in practice. Since the KL divergence of the approximate from the true posterior (\textit{i.e.,} first RHS term in Eq.~(\ref{eqn:1})) is non-negative,
we instead maximize the evidence lower bound (ELBO) $\mathcal{L}(\theta, \psi; \bm{X}^{(i)}_{1:M})$ as follows:
\begin{equation}\label{eq:elbo}
\begin{split}
    \mathcal{L}(\theta, \phi; \bm{X}^{(i)}_{1:M}) = \mathbb{E}_{q_{\phi}(\bm{z}|\bm{X}^{(i)}_{1:M})} [\log p_{\theta}(\bm{X}^{(i)}_{1:M}| \bm{z})] \\
    - D_{\text{KL}}(q_{\phi}(\bm{z} | \bm{X}^{(i)}_{1:M}) ||p_{\theta}(\bm{z})),
\end{split}
\end{equation}
where $\{q_{\phi_m}(\bm{z}|\bm{X})\}_{m=1}^M$ and $\{p_{\theta_m}(\bm{X}|\bm{z})\}_{m=1}^M$ are the $M$ probabilistic encoders and decoders, respectively. For notation brevity, we will omit the sample index $(i)$ hereafter.
An overview of the multimodal VAE is shown in Fig.~\ref{fig:graphical_abstract}.
However, in a multimodal scenario, maximizing the above ELBO objective requires the knowledge of the true joint posterior $p_{\theta}(\bm{z}|\bm{X}_{1:M})$, which is unknown in practice. To tackle this issue, previous explorations of multimodal VAEs approximate the true joint posterior by aggregating the unimodal inference distributions with a proper function $f_{\text{aggr}}(\cdot)$: 
\begin{equation}
\Tilde{q}(\bm{z}|\bm{X}_{1:M})=f_{\text{aggr}}(\{q_{\phi_m}\}_{m=1}^M),   \nonumber
\end{equation}
where $\Tilde{q}(\bm{z}|\bm{X}_{1:M})$ denotes the approximate joint posterior.
Some popular choices of $f$ are PoE~\cite{wu2018multimodal}, MoE~\cite{shi2019variational}, or a combination of both~\cite[MoPoE;][]{sutter2021generalized}. 
Mathematically, the approximate joint posterior $\Tilde{q}(\bm{z}|\bm{X}_{1:M})$ by PoE and MoE can be summarized as 
\begin{equation} \nonumber
    \Tilde{q}(\bm{z}|\bm{X}_{1:M}) = \begin{cases}
       \frac{1}{Z} \prod\limits_{m=1}^M q_{\phi_m}(\bm{z}|\bm{x}_m), \ \text{PoE}, \\
       \frac{1}{M} \sum\limits_{m=1}^M  q_{\phi_m}(\bm{z}|\bm{x}_m), \ \text{MoE}, 
    \end{cases}
\end{equation}
where $Z$ is the normalizer function that ensures the approximate posterior by PoE is a valid probability measure. 

\begin{figure*}[!t]
    \centering
    \includegraphics[width=0.9\textwidth]{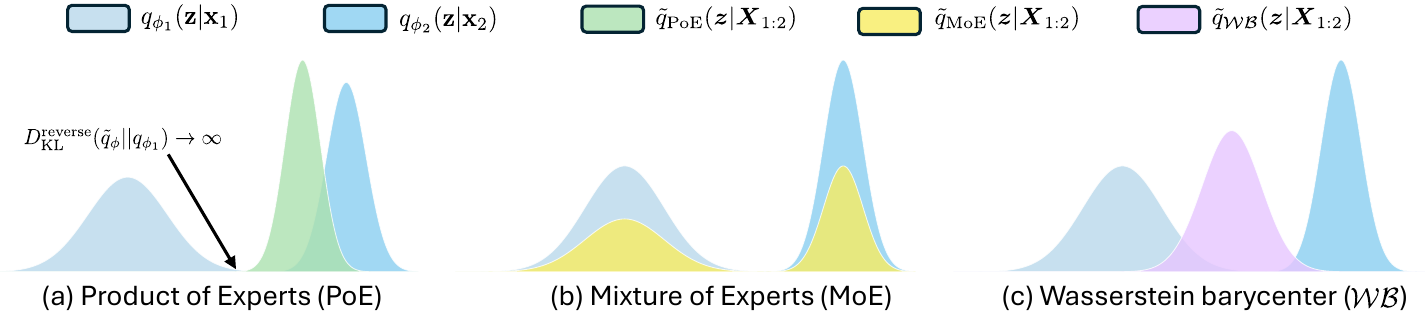}
    \caption{Comparison of methods for aggregating the unimodal inference distributions ($\{q_{\phi_j}\}_{j=1}^M$) to approximate the joint posterior ($\Tilde{q}_\phi$): (a) PoE, (b) MoE, and (c) the proposed Wasserstein barycenter. In this illustrative example, we use two 1-dimensional Gaussian modalities ($M=2$) for a proof of concept. }
    \label{fig:1D_example}
\end{figure*}

\subsection{Multimodal VAE: a Barycentric View}
The barycenter of distribution is defined as a central distribution of a set of distributions that minimizes the sum of divergences to all other distributions in the set. 
For a set of probability distributions $\{P_1, \cdots, P_M\}$ with associated weights $\{\lambda_1, \cdots, \lambda_M\}$,
the barycenter minimizes the weighted sum of some divergences $d(\cdot, \cdot)$ from the barycenter distribution $P_{\mathcal{B}}$ to each of the given distributions:
\begin{equation}\label{eq:barycenter}\nonumber
    P_{\mathcal{B}} = \argmin\limits_{P} \sum_{m=1}^M \lambda_m d(P_m, P), \quad \sum_{m=1}^M \lambda_m=1.
\end{equation}
\begin{lemma}\label{prop:bicentric_view}
   In the context of multimodal VAE, we seek to find a barycenter $\Tilde{q}(\bm{z} | \bm{X}_{1:M})$ that can aggregate the unimodal inference distributions $\{q_{\phi_m}(\bm{z}|\bm{x}_m)\}_{m=1}^M$ to approximate the true joint posterior $p_{\theta}(\bm{z} | \bm{X}_{1:M})$:  
\begin{equation}\label{eq:barycenter_vae}
\begin{split}
    \Tilde{q}  &=\argmin\limits_{q} \sum_{m=1}^M \lambda_m d\left(q_{\phi_m}, q \right), \ \sum_{m=1}^{M} \lambda_m = 1.
\end{split}
\end{equation} 
\end{lemma}
Note that, for notation brevity, we abbreviate $q_{\phi_m}(\bm{z} | \bm{x}_m)$ and $\Tilde{q}(\bm{z} | \bm{X}_{1:M})$ as $q_{\phi_m}$ and $\Tilde{q}$, respectively. 
Instead of directly minimizing the divergence between $q_{\phi}(\bm{z} | \bm{X}_{1:M})$ and $p_{\theta}(\bm{z})$ over trainable parameters $\phi=\{\phi_1, \cdots, \phi_M \}$ as formulated in Eq.~(\ref{eq:elbo}) and prior multimodal VAEs~\cite{wu2018multimodal,shi2019variational,sutter2020multimodal,sutter2021generalized}, Lemma~\ref{prop:bicentric_view} suggests that this involves a bilevel optimization. For the lower-level optimization (\textit{i.e.,} Eq.~(\ref{eq:barycenter_vae})),
we determine a barycenter $\Tilde{q}(\bm{z} | \bm{X}_{1:M})$, which is equivalent to applying an aggregation function $f_{\text{aggr}}$ to combine the unimodal inference distributions. We then push $\Tilde{q}(\bm{z} | \bm{X}_{1:M})$ towards $p_{\theta}(\bm{z})$ by minimizing their divergence over trainable parameters $\phi=\{\phi_m\}_{m=1}^M$ (upper-level optimization; Eq.~(\ref{eq:elbo})). At first glance, this formulation is counterintuitive, as it complicates the formulation and optimization, whereas in-depth analysis reveals its theoretically intriguing properties.

\begin{prop}\label{lemma:1}
    For any divergence measure $d(q_{\phi_m}, \cdot)$ that is convex on $q_{\phi_m}$, the resultant barycenter by minimizing Eq.~(\ref{eq:barycenter_vae}) guarantees a valid ELBO on the marginal log-likelihood $p_\theta(\bm{X}_{1:M})$ and a scalable inference. This is because of Jensen's inequality:
        \begin{equation}\label{eq:remark:1}
         d\left(\sum_{m=1}^M \lambda_m q_{\phi_m}, q \right) \leq \sum_{m=1}^M \lambda_m d(q_{\phi_m}, q)
    \end{equation}
\end{prop}

For a complete proof of Proposition~\ref{lemma:1}, please see \textbf{Appendix A.1}.
The LHS in Eq.~(\ref{eq:remark:1}) defines a scalable inference, as the naive implementation on the RHS requires $2^M$ inference networks to handle arbitrary combination of input modalities.
Although Proposition~\ref{lemma:1} has been considered in some prior works from different perspectives~\cite{shi2019variational,sutter2021generalized}, they are limited to the case of KL divergence (see Theorem~\ref{thm:1}). In contrast, our barycentric view extends them to a more general case whenever $d(q_{\phi_m}, q)$ is convex to $q_{\phi_m}$, which enables it to analyze the properties of a more flexible choice of divergence measures (\textit{e.g.,} $f$-divergence, 2-Wasserstein distance, Gromov-Wasserstein distance, etc).

\begin{theorem}\label{thm:1}
    Considering KL divergence $D_{\text{KL}}(\cdot || \cdot)$ as the divergence measure $d(\cdot, \cdot)$, PoE and MoE are the barycenters yielded by optimizing the reverse and forward KL divergence, respectively:
    \begin{align}
        \Tilde{q}_{\text{PoE}} &= \argmin\limits_{q} \frac{1}{Z} \sum_{m=1}^M D^{\text{reverse}}_{\text{KL}}(q || q_{\phi_m}), \nonumber \\
        \Tilde{q}_{\text{MoE}} &=\argmin\limits_{q} \frac{1}{M} \sum_{m=1}^M D^{\text{forward}}_{\text{KL}}(q_{\phi_m} || q). \nonumber\label{eq:bary_moe}
    \end{align}
\end{theorem}

The proof of Theorem~\ref{thm:1} is in \textbf{Appendix A.2}. In information theory, it is customary to define KL divergence as relative entropy (due to its asymmetry), with the form used in PoE and MoE in Theorem~\ref{thm:1} being the 
exclusive (reverse) and inclusive (forward) KL divergence~\cite{cover1999elements,murphy2012machine}. 
Theorem~\ref{thm:1} immediately provides an information-theoretic view of PoE and MoE: they are two variants resulting from the inherent asymmetry of KL divergence. this provides us with an information-theoretic tool to analyze the properties of PoE and MoE in multimodal VAE. 

\begin{remark}\label{remark:3}
    PoE is zero-forcing, encouraging $\Tilde{q}(\bm{z} | \bm{X}_{1:M})$ to be zero where $q_{\phi_m}(\bm{z}|\bm{x}_m)$ is zero, which makes it biased towards certain modalities. In contrast, MoE is mass-covering, ensuring that there is mass under $\Tilde{q}(\bm{z} | \bm{X}_{1:M})$ wherever there is mass under $q_{\phi_m}(\bm{z}|\bm{x}_m)$.
\end{remark}

Remark~\ref{remark:3} is due to the intrinsic properties of forward and reverse KL divergence~\cite{minka2005divergence,turner2011two}. Though it is well known that PoE results in a sharper distribution that concentrates on one of the modalities, whereas MoE does not produce a distribution sharper than any individual expert due to the nature of the mixture, Remark~\ref{remark:3} provides an information-theoretic interpretation. 
We demonstrate this by considering an example with two modalities, as shown in Fig.~\ref{fig:1D_example}.
When there is zero mass under $q_{\phi_1}$ and nonzero mass under $\Tilde{q}_{\text{PoE}}$, the reverse KL divergence is almost infinity: $D_{\text{KL}}^{\text{reverse}}(\Tilde{q}_{\text{PoE}}||q_{\phi_1}) \rightarrow \infty$, which pushes $\Tilde{q}_{\text{PoE}}$ toward $q_{\phi_{2}}$ (see Fig.~\ref{fig:1D_example}a). In contrast, since the forward KL divergence penalizes $\log q_{\phi_m}(\bm{z}|\bm{x}_m) - \log \Tilde{q}(\bm{z} | \bm{X}_{1:M})$, it ensures that $\Tilde{q}$ has mass covered wherever this is mass under $q_{\phi_m}$ (see Fig.~\ref{fig:1D_example}b). 
 
However, the forward and reverse KL divergence does not define a metric space for probability measures because it is asymmetric and unbounded. One notable example is that solving Eq.~(\ref{eq:barycenter_vae}) does not guarantee a valid probability measure in the case of PoE (see \textbf{Appendix A.2}).
This motivates us to find a barycenter defined in the probability metric space. Below, we explore the barycenter defined in the 2-Wasserstein space, known as the Wasserstein barycenter. 

\subsection{Multimodal VAE from Wasserstein Barycenter}
Here, we provide a roadmap to derive the proposed Wasserstein barycenter VAE ($\mathcal{WB}$-VAE) for multimodal representation learning.
Following the convention in Eq.~(\ref{eq:barycenter}), Wasserstein barycenter ($\mathcal{WB}$) is defined by minimizing the squared 2-Wasserstein distance $\mathcal{W}_2^2(\cdot, \cdot)$:
\begin{equation}\label{eq:wass_barycenter}
    P_{\mathcal{WB}} = \argmin\limits_{P} \sum_{m=1}^M \lambda_m \mathcal{W}_2^2(P_m, P), \quad \sum_{m=1}^M \lambda_m=1. \nonumber
\end{equation}
Since the 2-Wasserstein distance is symmetric, the order of distributions in $\mathcal{W}(\cdot, \cdot)$ does not matter. In the context of multimodal VAE, the approximate posterior resulting from optimizing the squared 2-Wasserstein distance is 
\begin{equation}\label{eq:bures_wass_vae}
\begin{split}
    \Tilde{q}_{\mathcal{WB}}&  =\argmin\limits_{q} \sum_{m=1}^M \lambda_m \mathcal{W}_2^2(q_{\phi_m}, q), 
    \ \sum_{m=1}^{M} \lambda_m = 1. \nonumber
\end{split}
\end{equation}
Unlike the KL divergence used in the case of PoE and MoE, which focuses on pointwise differences, the 2-Wasserstein distance better preserves the geometry of the unimodal inference distributions. Accordingly, interpolating in the Wasserstein space (i.e., a geodesic space) can have a meaningful transition from unimodal distributions to the joint posterior, especially when the unimodal distributions have different shapes or supports~\cite{ambrosio2008gradient}. Therefore, different choices of weights associated with unimodal distributions (i.e., $\{ \lambda_1, \cdots, \lambda_M \}$) may lead to a joint posterior that maintains diverse shapes and structures of unimodal distributions. However, in the context of multimodal VAEs, it is challenging to determine $\{ \lambda_1, \cdots, \lambda_M \}$, as we only have the marginal unimodal distributions. Similar to the case of PoE and MoE, it is typically safe to set $\lambda_m=1/M, \ \forall m$. 

\subsubsection{Bures-Wasserstein barycenter.} Wasserstein barycenter typically incurs the significant computational cost associated with the 2-Wasserstein distance. However, in the case of Gaussian distributions, as are typically assumed in VAEs, the Gaussian Wasserstein barycenter (\textit{i.e.,} the so-called Bures-Wasserstein barycenter~\cite{agueh2011barycenters}) can be obtained by solving a fixed-point equation~\cite{knott1994generalization,agueh2011barycenters}. 

Considering the unimodal inference distributions $\{q_{\phi_m}\}_{m=1}^{M}$ are $d$-dimensional multivariate Gaussian $\{ \mathcal{N}(\bm{\mu}_m, \bm{\Sigma}_m) \}_{m=1}^M$, with $\bm{\mu}_m \in \mathbb{R}^d$ and $\bm{\Sigma}_m \in \mathbb{R}^{d \times d}$ being the associated mean and covariance of $q_{\phi_m}$, the resultant Bures-Wasserstein barycenter turns out to be Gaussian-distributed, \textit{i.e.,} $\Tilde{q}_{\mathcal{WB}}(\bm{z} | \bm{X}_{1:M})  \sim \mathcal{N}(\bm{\Tilde{u}}, \bm{\Tilde{\Sigma}})$:
\begin{equation}\label{eq:bures_wass_bc}
    \bm{\Tilde{\mu}} = \sum_{m=1}^{M} \lambda_m \bm{\mu}_m, \ \bm{\Tilde{\Sigma}} =\sum_{m=1}^{M} \lambda_m (\bm{\Tilde{\Sigma}}^{1/2} \bm{\Sigma}_m  \bm{\Tilde{\Sigma}}^{1/2})^{1/2},
\end{equation}
where the covariance $\bm{\Tilde{\Sigma}}$ is obtained by solving the fix-point equation in Eq.~(\ref{eq:bures_wass_bc}). However, Eq.~(\ref{eq:bures_wass_bc}) can be further simplified by considering $q_{\phi_m}(\bm{z} | \bm{x}_m)$ an isotropic Gaussian with a diagonal covariance $\mathcal{N}(\bm{\mu}_m, \bm{\sigma}_m^2 \bm{I})$ with $\bm{\mu}_m, \bm{\sigma}_m \in \mathbb{R}^d$ and $\bm{I} \in \mathbb{R}^{d \times d}$. This is also typically assumed in most VAEs~\cite{vae}. 

\begin{remark}\label{remark:bures_wass}
    In the isotropic Gaussian case, Eq.~(\ref{eq:bures_wass_bc}) can be solved analytically dimension by dimension:
    \begin{equation}
    \Tilde{\bm{\mu}} = \sum_{m=1}^{M} \lambda_m \bm{\mu}_m, \ \Tilde{\bm{\sigma}} = \sum_{m=1}^{M} \lambda_m \bm{\sigma}_m .
\end{equation}
\end{remark}

Remark~\ref{remark:bures_wass} is because the optimal transport map from one Gaussian to another is a linear map~\cite{knott1994generalization,agueh2011barycenters}, with which the squared 2-Wasserstein distance can be solved analytically (for details, please see~\textbf{Appendix A.3}). As suggested by Lemma~\ref{lemma:1}, the Bures-Wasserstein barycenter can be viewed as minimizing the 2-Wasserstein distance to a mixture of distributions.

\subsubsection{Mixture of Wasserstein barycenter.}
The approximate joint distribution derived from solving the Wasserstein barycenter strikes a balance between zero-forcing  (bias) and mass-covering (variance), resulting in a distribution that is sharper than half of the unimodal inference distributions (see Fig.~\ref{fig:1D_example}c). However, there is an inherent trade-off between zero-forcing and mass-covering~\cite{murphy2012machine}. 
Similar to MoPoE-VAE~\cite{sutter2021generalized}, we consider a variant of $\mathcal{WB}$-VAE by constructing a mixture of Wasserstein barycenter, termed $\mathcal{MWB}$-VAE.

\begin{remark}
    The mixture of Wasserstein barycenter with unimodal inference distributions is still a barycenter. Considering the powerset of $M$ modalities $\mathcal{P}_M(\bm{X})$, which consists of $2^M$ different combinations, the mixture of Wasserstein barycenter is given as
    \begin{equation}\nonumber
    \begin{split}
         &\quad \Tilde{q}_{\mathcal{MWB}} = \argmin_{q} \sum_{\bm{X}_k 
     \in \mathcal{P}_M(\bm{X})}\lambda_k  D_{\text{KL}}(\Tilde{q}_{\mathcal{WB}} || q) \\
     &\text{subject to} \quad \Tilde{q}_{\mathcal{WB}}  =\argmin\limits_{q} \sum_{\bm{x}_j \in \bm{X}_k} \lambda_j \mathcal{W}_2^2(q_{\phi_j}, q)
    \end{split}
    \end{equation} 
\end{remark}
Though this is a bilevel optimization problem, the solution is analytical since both the lower-level and upper-level optimization problems can be solved analytically. The solution is also optimal due to the convexity of both forward KL divergence and 2-Wasserstein distance. By applying the same mechanism, we can also derive MoPoE~\cite{sutter2021generalized} as a barycenter, whereas the solution is not guaranteed to be optimal since the solution to the lower-level (PoE) case is not a global optimum in general.

\begin{figure*}[!t]
    \centering
    \includegraphics[width=0.8\textwidth]{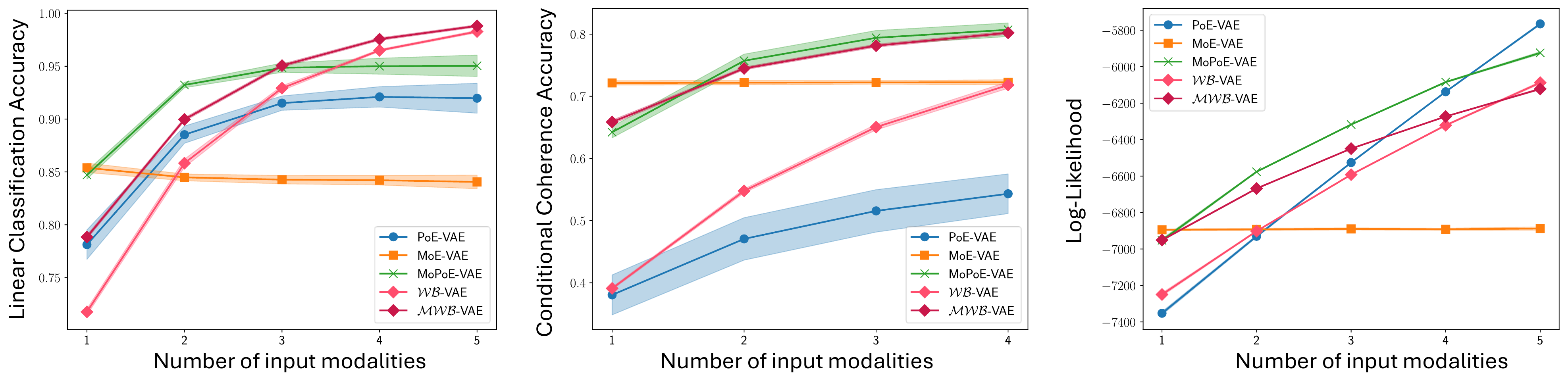}
    \caption{Quantitative results on PolyMNIST as a function of the number of input modalities, averaged over all subsets of modalities of the respective size. \textbf{Left}: Linear classification accuracy of digits given the latent representation. \textbf{Center}: Coherence of conditionally generated samples that do not include input modalities. \textbf{Right}: Log-Likelihood of all generated modalities. }
    \label{fig:poly_mnist}
\end{figure*}
\vspace{-0.2cm}

\begin{table*}[!t]
\caption{Linear classification accuracy of latent representations for MNIST-SVHN-TEXT. We evaluated all possible combinations of modalities $\bm{X}_k$. We reported the means ($\pm$ standard deviations) over 5 runs, where the best performance is highlighted with \textbf{bold}. The abbreviations of different modalities in this table are as follows: M: MNIST; S: SVHN; T: Text. }
\vspace{-0.2cm}
\label{tab:mst_representation_classification}
\begin{center}
    \begin{threeparttable}
     \resizebox{0.69\textwidth}{!}{
    \begin{tabular}{l|cccccccc}
         \toprule
        Model &  M & S & T & M,S & M,T & S,T & M,S,T & Avg.\\
        \midrule
        PoE-VAE & 0.90$\pm$0.01 & 0.44$\pm$0.01 & 0.85$\pm$0.10 & 0.89 $\pm$0.01 & 0.97$\pm$0.02 & 0.81$\pm$0.09 & 0.96$\pm$0.02 & 0.83 \\
        MoE-VAE & 0.95$\pm$0.01 & 0.79$\pm$0.05 & 0.99$\pm$0.01 & 0.87$\pm$0.03 & 0.93$\pm$0.03 & $0.84$$\pm$0.04 & 0.86$\pm$0.03 & 0.89 \\
        MoPoE-VAE & 0.95$\pm$0.01 &0.80$\pm$0.03 &0.99$\pm$0.01 & 0.97$\pm$0.01 & 0.98$\pm$0.01 & 0.99$\pm$0.01 & 0.98$\pm$0.01 & 0.95 \\
        \midrule
        \rowcolor{lightgray!20}
        $\mathcal{WB}$-VAE & 0.91$\pm$0.03 & 0.44$\pm$0.02 & 1.00$\pm$0.00 & 0.89$\pm$0.00 & 0.99$\pm$0.02 & 0.99$\pm$0.01 &  0.99$\pm$0.00 & 0.89 \\
        \rowcolor{lightgray!20}
         $\mathcal{MWB}$-VAE & \textbf{0.97}$\pm$0.00 &\textbf{0.83}$\pm$0.01 &\textbf{1.00}$\pm$0.00 & \textbf{0.99}$\pm$0.00 & \textbf{1.00}$\pm$0.00 & \textbf{1.00}$\pm$0.00 & \textbf{1.00}$\pm$0.00 & \textbf{0.97}$^*$ \\
         \bottomrule
    \end{tabular}}
\end{threeparttable}
\end{center}
\end{table*}

\begin{table}[!t]
\caption{Conditional generation coherence for MNIST-SVHN-TEXT. The modality above the horizontal line indicates the one generated based on the subsets $\bm{X}_k$ listed below. We reported the mean values over 5 runs, where the best performance is highlighted with $\textbf{bold}$.}
\vspace{-0.2cm}
\label{tab:mst_generation_coherence}
\begin{threeparttable}
\begin{center}
\resizebox{0.95\columnwidth}{!}{
    \begin{tabular}{l|cccccccccc}
        \toprule
        &  \multicolumn{3}{c}{M} & \multicolumn{3}{c}{S} & \multicolumn{3}{c}{T} \\
        \cmidrule(l){2-4} \cmidrule(l){5-7} \cmidrule(l){8-10}
        Model &  S & T & S,T & M & T & M,T & M & S & M,S & Avg. \\
        \midrule
        PoE-VAE & 0.24 & 0.20 & 0.32 & \textbf{0.43} & 0.30 & \textbf{0.75} & 0.28 & 0.17 & 0.29 & 0.32 \\
        MoE-VAE  & 0.75 & 0.99 & 0.87 & 0.31 & 0.30 & 0.30 & 0.96 & 0.76 & 0.84 &0.68 \\
        MoPoE-VAE & 0.74 & 0.99 & 0.94 & 0.36 & 0.34 & 0.37 & 0.96 & 0.76 & 0.93 &0.71 \\
        \midrule
        \rowcolor{lightgray!20}
        $\mathcal{WB}$-VAE  & 0.12 & 0.51 & 0.57 & 0.28 & \textbf{0.39} & 0.53 & 0.52 &  0.18 & 0.57 & 0.41 \\
        \rowcolor{lightgray!20}
        $\mathcal{MWB}$-VAE & \textbf{0.82} & \textbf{1.00} & \textbf{0.99} & 0.36 & 0.35 & 0.39 & \textbf{0.97} & \textbf{0.84} & \textbf{0.99} & \textbf{0.75}$^*$ \\
        \bottomrule
    \end{tabular}}
\end{center}
\end{threeparttable}
\end{table}

\begin{table*}[!t]
\caption{The log-likelihoods of the joint generative model conditioned on the approximate joint posterior $\Tilde{q}(\bm{z}|\bm{X}_{1:M})$ on the MNIST-SVHN-TEXT test set. The means ($\pm$ standard deviations) of 5 runs were reported. We highlight the best performance in \textbf{bold}, according to the first three decimals.
($\bm{x}_M$: MNIST; $\bm{x}_S$: SVHN; $\bm{x}_T$: Text; $\bm{X} = (\bm{x}_M, \bm{x}_S,\bm{x}_T)$).}
\vspace{-0.2cm}
\label{tab:mst_likelihoods}
\begin{center}
\resizebox{0.68\textwidth}{!}{
\begin{tabular}{l|ccccccc}
    \toprule
    Model & $\bm{X}$ & $\bm{X}|\bm{x}_M$ & $\bm{X}|\bm{x}_S$ & $\bm{X}|\bm{x}_T$ & $\bm{X}|\bm{x}_M,\bm{x}_S$ & $\bm{X}|\bm{x}_M,\bm{x}_T$ & $\bm{X}|\bm{x}_S,\bm{x}_T$ \\
    \midrule
    PoE-VAE & -1790$\pm$3.3 & -2090$\pm$3.8 & -1895$\pm$0.2 & -2133$\pm$6.9 & -1825$\pm$2.6 & -2050$\pm$2.6 & \textbf{-1855}$\pm$0.3 \\
    MoE-VAE & -1941$\pm$5.7 & \textbf{-1987}$\pm$1.5 & -\textbf{1857}$\pm$12 & \textbf{-2018}$\pm$1.6 & -1912$\pm$7.3 & -2002$\pm$1.2 & -1925$\pm$7.7 \\
    MoPoE-VAE & -1819$\pm$5.7 & -1991$\pm$2.9 & \textbf{-1858}$\pm$6.2 & -2024$\pm$2.6 & -1822$\pm$5.0 & \textbf{-1987}$\pm$3.1 & \textbf{-1850}$\pm$5.8 \\
    \midrule
    \rowcolor{lightgray!20}
    $\mathcal{WB}$-VAE & \textbf{-1785}$\pm$7.4 & -2072$\pm$13 & -1889$\pm$7.4 &-2126$\pm$12 & \textbf{-1814}$\pm$7.5 & -2033$\pm$7.1 & \textbf{-1856}$\pm$4.7 \\
    \rowcolor{lightgray!20}
     $\mathcal{MWB}$-VAE & -1890$\pm$1.7 & -2000$\pm$1.4 & \textbf{-1856}$\pm$3.4 & -2036$\pm$0.4 & -1825$\pm$1.6 & \textbf{-1988}$\pm$1.4 & \textbf{-1853}$\pm$2.2 \\
     \bottomrule
\end{tabular}}
\end{center}
\end{table*}

\section{Experiments}
\subsubsection{Dataset.}
We conducted comparative experiments on three multimodal benchmark datasets: i) PolyMNIST with five simplified modalities, ii) the trimodal MNIST-SVHN-TEXT, and iii) the challenging bimodal CelebA dataset. PolyMNIST was generated by combining each MNIST digit~\cite{lecun-mnisthandwrittendigit-2010} with $28\times28$ random crops from five distinct background images, as described in~\cite{sutter2021generalized}. This process generated five different modalities, each consisting of an MNIST digit overlaid on a background crop.
The MNIST-SVHN-TEXT dataset was introduced by \cite{sutter2020multimodal}, which consists of three modalities: MNIST digit~\cite{lecun-mnisthandwrittendigit-2010}, text, and SVHN~\cite{netzer2011reading}. The MNIST digit and text are two clean modalities, whereas SVHN is comprised of noisy images. Folliwing~\cite{sutter2021generalized}, 20 triples were generated per set using a many-to-many mapping.  The bimodal CelebA includes human face images as well as text describing the face attributes~\cite{liu2015deep}. This dataset is challenging because the text modality focuses on the attributes present in a face image. If an attribute is absent, it is omitted from the corresponding text \cite{sutter2020multimodal}.

\noindent\textbf{Baseline methods.} We compared the proposed method to three state-of-the-art multimodal VAEs, including PoE-VAE~\cite{wu2018multimodal}, MoE-VAE~\cite{shi2019variational}, and MoPoE-VAE~\cite{sutter2021generalized}.

\noindent\textbf{Evaluation metric.} 
Following previous literature in~\citet{wu2018multimodal,shi2019variational,sutter2021generalized}, several tasks were conducted to evaluate the performance of the multimodal VAEs. First, a linear classifier was used to assess the quality of the learned latent representations. Second, the coherence of generated samples was evaluated using pre-trained classifiers. Third, the approximate joint posterior was measured by calculating the log-likelihoods on the test set. 

\noindent\textbf{Implementation details.} For a fair comparison, we followed the experimental settings in previous literature~\cite{shi2019variational,sutter2021generalized}. In particular, we employed the same network architecture as in~\cite{shi2019variational,sutter2021generalized}. For more implementation details (e.g., hyperparameter configurations), we kindly direct the readers to \textbf{Appendix B}. All experiments were performed on a Nvidia-A100 GPU with 40G memory. 

\begin{figure*}[!t]
    \centering
    \includegraphics[width=0.65\textwidth]{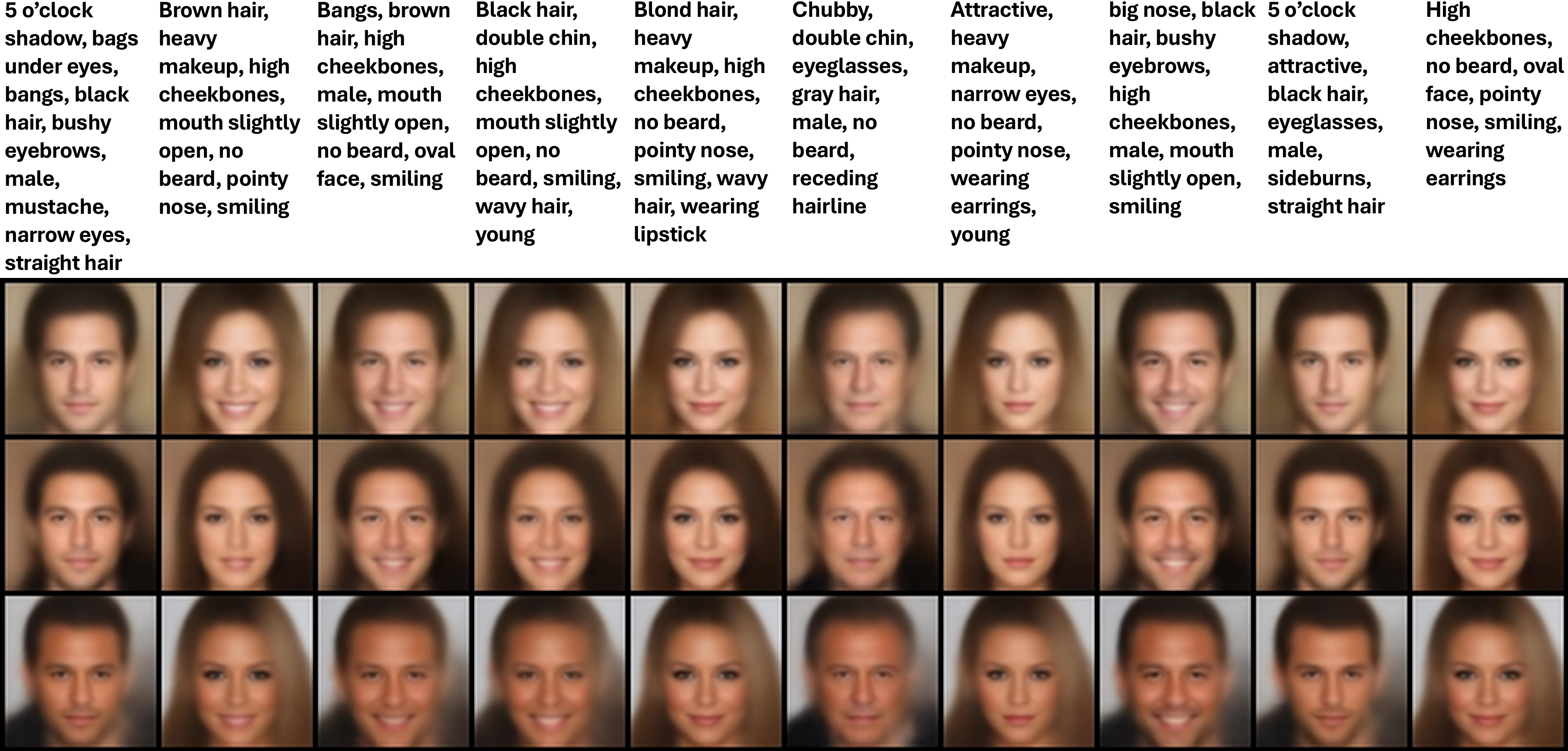}
    \caption{Conditionally generated images given the text on top of each column on bimodal CelebA using $\mathcal{MWB}$-VAE.}
    \label{fig:vis_celeba}
\end{figure*}

\begin{table}[!t]
\caption{Classification accuracy based on latent representation and conditionally generated coherence on the bimodal CelebA dataset. We report the mean average precision over all attributes (I: Image; T: Text; Joint: I and T). The best performance is highlighted in \textbf{bold}.}
\vspace{-0.2cm}
\label{tab:celeba_classification}
\begin{threeparttable}
\begin{center}
\resizebox{0.85\columnwidth}{!}{
\begin{tabular}{l|cccccc}
    \toprule
    & \multicolumn{3}{c}{Latent Representation} & \multicolumn{2}{c}{Generation} \\
    \cmidrule(l){2-4} \cmidrule(l)                                                      {5-6}
    Model & I & T & Joint & I $\rightarrow$ T & T $\rightarrow$ I & Avg. \\
    \midrule
    PoE-VAE & 0.30 & 0.31 & 0.32 & 0.26 & 0.33 & 0.30  \\
    MoE-VAE & 0.35 & 0.38 & 0.35 & 0.14 & 0.41  & 0.33 \\
    MoPoE-VAE & \textbf{0.40} & 0.39 & 0.39 & 0.15 & \textbf{0.43} & 0.35 \\
    \midrule
    \rowcolor{lightgray!20}
    $\mathcal{WB}$-VAE & 0.34  & 0.38 & 0.40 &  0.29 & 0.40 & 0.36 \\
    \rowcolor{lightgray!20}
    $\mathcal{MWB}$-VAE & 0.37  & \textbf{0.44} & \textbf{0.44} & \textbf{0.34} & \textbf{0.43} & \textbf{0.40}$^*$  \\
    \bottomrule
\end{tabular}}
\end{center}
\vspace{-0.2cm}
\end{threeparttable}
\end{table}

\section{Results}

\subsubsection{PolyMNIST results.} 
The PolyMNIST dataset is unique in that it contains more than three modalities, enabling us to explore how different methods perform as the number of input modalities increases (see Fig.~\ref{fig:poly_mnist}).
Notably, the proposed $\mathcal{WB}$-VAE and $\mathcal{MWB}$-VAE showed an approximately linear relationship between all the performance metrics and the number of input modalities. This is because adding more modalities is analogous to interpolating in Wasserstein space, which generally results in a smooth transition within the probability space~\cite{ambrosio2008gradient}.
This was particularly true for the linear classification task, where the performance of other baseline methods was typically saturated after reaching a certain number of modalities (e.g., $M>3$ in Fig.~\ref{fig:poly_mnist} \textbf{Left}). As a consequence,  $\mathcal{WB}$-VAE and $\mathcal{MWB}$-VAE showed superior performance in terms of linear classification accuracy compared to all baseline methods, particularly when the number of input modalities increases.  
Similar trends were also observed in the conditional generation task (Fig.~\ref{fig:poly_mnist} \textbf{Center}), where the generation coherence of $\mathcal{WB}$-VAE increased as the number of input modalities increased. Although $\mathcal{WB}$-VAE outperformed PoE-VAE, it did not surpass MoE-VAE, but it struck the balance between them, as there is an inherent trade-off between mass-covering and zero-forcing. As a consequence, $\mathcal{MWB}$-VAE can easily outperform MoE-VAE and achieve similar performance as MoPoE-VAE in the conditional generation task. As suggested by~\citet{sutter2021generalized}, there is a trade-off between generation coherence and the log-likelihood. Consequently, the PoE-VAE achieved the highest log-likelihood. Although $\mathcal{WB}$-VAE and $\mathcal{MWB}$-VAE did not surpass PoE-VAE in log-likelihood, their log-likelihoods were on par with MoPoE-VAE.

\noindent \textbf{MNIST-SVHN-TEXT results.} 
As shown in Tables~\ref{tab:mst_representation_classification} and~\ref{tab:mst_generation_coherence}, the proposed $\mathcal{MWB}$-VAE demonstrated superior performance compared to other state-of-the-art multimodal VAEs in terms of the quality of learned latent representations and generation coherence. In addition, our $\mathcal{WB}$-VAE outperformed PoE-VAE regarding the linear classification accuracy using the learned latent representations and was on par with PoE-VAE regarding generation coherence. Although there is an inherent trade-off between generation coherence and log-likelihood, the log-likelihood of our $\mathcal{WB}$-VAE and $\mathcal{MWB}$-VAE were on par with the other state-of-the-art methods.
This suggests that the proposed method can approximate the joint posterior well.

\noindent \textbf{CelebA results.}
As shown in Table~\ref{tab:celeba_classification}, the proposed $\mathcal{WB}$-VAE outperformed PoE-VAE as well as competed favorably and even better than MoE-VAE in both latent representation and generation on the challenging bimodal CelebA dataset. Likewise, $\mathcal{MWB}$-VAE outperformed MoPoE-VAE in most scenarios, with the exception of latent representation classification when using image as the input modality. As consistent with the trends observed in the previous two datasets, the latent representation classification accuracy of $\mathcal{WB}$-VAE increased as more modalities were present, similar to PoE-VAE. In contrast, the classification accuracy of MoE-VAE decreased when more modalities were given.
Remarkably, both $\mathcal{WB}$-VAE and $\mathcal{MWB}$-VAE achieved good performance for the most challenging image-to-text generation task, outperforming the second-best method by 11.5\% and 30.8\%, respectively. $\mathcal{MWB}$-VAE also achieved good performance in text-to-image conditional generation (see Fig.~\ref{fig:vis_celeba}), where $\mathcal{MWB}$ learned good representations of different attributes well (\textit{e.g.,} "smiling," "hairstyles," etc).

\section{Conclusion}
In this work, we introduced a barycentric perspective on previous multimodal VAEs, offering a theoretical and unified formulation. This approach allows for explorations of various aggregation functions in the regime of multimodal VAEs. Leveraging this barycentric formulation, we proposed a $\mathcal{WB}$-VAE, which uses the Wasserstein barycenter as an aggregation function that better preserves the geometry of unimodal distributions. Experimental results showed the effectiveness of the proposed $\mathcal{WB}$-VAE when compared to other state-of-the-art multimodal VAEs. We hope our new perspective will stimulate the exploration of other aggregation functions for multimodal VAEs in future work.

\section{Acknowledgments}
This work was partially supported by NIH grant
R01-AG067103. 
Computations were performed using the resources
of the Washington University Research Computing and Informatics
Facility, which were partially funded by NIH grants S10OD025200,
1S10RR022984-01A1 and 1S10OD018091-01.


\bibliography{aaai25}

\appendix
\renewcommand{\thefigure}{S\arabic{figure}}
\renewcommand{\thetable}{S\arabic{table}}

\section*{Supplementary Material}

\section{A \quad Proofs}\label{sec:proofs}

\subsection{A.1 \quad Proof of Proposition 1}
\begin{proof}
    The proof of Proposition~\ref{lemma:1} can be carried out by showing that ELBO is the lower bound of the log-likelihood:
    \begin{equation}
        \log p_{\theta}(\bm{X}_{1:M}) \geq \mathcal{L}(\theta, \phi; \bm{X}_{1:M}), \nonumber
    \end{equation}
or, equivalently
\begin{equation}
    D_{\text{KL}}(\Tilde{q}(\bm{z} | \bm{X}_{1:M}) ||p_{\theta}(\bm{z}|\bm{X}_{1:M})) \geq 0. \nonumber
\end{equation}

Due to Jensen's inequality, for any divergence measure $d(q_m, \cdot)$ that is convex on $q_m$, we can minimize the convex combination of $d(q_m, \cdot)$ for the barycenter. Therefore, the resultant barycentric distribution can be abstracted as any arbitrary function $f(\cdot)$ of the weighted combination of the unimodal posteriors:
\begin{equation}
    \Tilde{q}(\bm{z} | \bm{X}_{1:M}) = f(\mathcal{M}(\{q_{\phi_m}\}_{m=1}^M)). \nonumber
\end{equation}
Here, in a more strict sense, $\sum_{m=1}^M \lambda_m q_{\phi_m}$ is abstracted as the mixture of distributions $\mathcal{M}(\{q_{\phi_m}\}_{m=1}^M)$. 

In the case of KL divergence as $d(\cdot, \cdot)$, it is obvious that $\Tilde{q}(\bm{z} | \bm{X}_{1:M})$ is reduced to the mixture of experts. Although in a more general definition of an arbitrary function $d(\cdot, \cdot)$ where the aggregation function $f$ can be more complex and may not be analytical, the result of such minimization is still a single distribution. One example is in the case of the squared 2-Wasserstein distance, where the resultant single distribution is obtained by minimizing the squared 2-Wasserstein distance to the mixture distribution $\mathcal{M}(\{q_{\phi_m}\}_{m=1}^M)$. 
Therefore, it is trivial that 
\begin{equation}
     D_{\text{KL}}(f(\mathcal{M}(\{q_{\phi_m}(\bm{z}|\bm{x}_m)\}_{m=1}^M) ||p_{\theta}(\bm{z}|\bm{X}_{1:M})) \geq 0. \nonumber
\end{equation}
However, there is no guarantee of a valid ELBO on the log-likelihood for any divergence measure $d(q_m, \cdot)$ that is non-convex on $q_m$. 
\end{proof}

\subsection{A.2 \quad Proof of Theorem 1}

\begin{proof}
Without loss of generality, we prove a more general case of Theorem~\ref{thm:1}, under the condition that $\sum_{m=1}^M \lambda_m = 1$ without assuming equal weights (i.e., $\lambda_m=1/M, \  \forall m$). For notation brevity, we omit the subscripts (\textit{i.e.,} $1:M$) in $\bm{X}_{1:M}$, denoting $\Tilde{q}(\bm{z}|\bm{X}_{1:M})$ as $\Tilde{q}(\bm{z}|\bm{X})$ hereafter.

\subsubsection{Product of Experts:} We first show that the product of experts (PoE) used in~\citet{wu2018multimodal} is a barycenter yielded by optimizing the weighted sum of the reverse KL divergences:
    \begin{align*}
        \Tilde{q}_{\text{PoE}} &= \argmin\limits_{q}  \sum_{m=1}^M \lambda_m D^{\text{reverse}}_{\text{KL}}(q || q_{\phi_m}) \\
        &= \argmin\limits_{q(\bm{z}|\bm{X})} \sum_{m=1}^M  \int q(\bm{z}|\bm{X}) \log \left[\frac{q(\bm{z}|\bm{X})}{q_{\phi_m}(\bm{z} | \bm{x}_m)} \right]^{\lambda_m} d\bm{z}\\
        &=\argmin\limits_{q(\bm{z}|\bm{X})} \int q(\bm{z}|\bm{X}) \log  \prod_{m=1}^M \left[\frac{q(\bm{z}|\bm{X})}{q_{\phi_m}(\bm{z}|\bm{x}_m)} \right]^{\lambda_m} d\bm{z} \\
        &= \argmin\limits_{q(\bm{z}|\bm{X})} \int q(\bm{z}|\bm{X}) \log  \frac{[q(\bm{z}|\bm{X})]^{\sum_{m=1}^M \lambda_m}}{\prod_{m=1}^M \left[q_{\phi_m}(\bm{z}|\bm{x}_m)\right]^{\lambda_m}}  d\bm{z} \\
        &= \argmin\limits_{q(\bm{z}|\bm{X})} D_{\text{KL}}\left(q(\bm{z}|\bm{X}) \Big\Vert \prod_{m=1}^M\left[q_{\phi_m}(\bm{z}|\bm{x}_m)\right]^{\lambda_m} \right) 
    \end{align*}
The KL divergence in the last line is minimized when $q(\bm{z}|\bm{X})=\prod_{m=1}^M\left[q_{\phi_m}(\bm{z}|\bm{x}_m)\right]^{\lambda_m}$, However, the resulting distribution $\Tilde{q}_{\text{PoE}}(\bm{z}|\bm{X}_{1:M})= \left[ \prod_{m=1}^M q_{\phi_m}(\bm{z}|\bm{x}_m) \right]^{\lambda_m}$ may not be a valid probability distribution without normalization. 
 Therefore,
we typically define the PoE as $\Tilde{q}_{\text{PoE}}(\bm{z}|\bm{X}_{1:M})= \frac{1}{Z} \prod_{m=1}^M q_{\phi_m}(\bm{z}|\bm{x}_m)$, with $Z$ being the normalizer to ensure the distribution yielded by PoE a valid probability distribution: $Z=\int \prod_{m=1}^M \left[q_{\phi_m}(\bm{z}|\bm{x}_m)\right]^{\lambda_m} d\bm{z}$. 

\subsubsection{Mixture of Experts:} Similarly, we can show that the mixture of experts (MoE) used in~\citet{shi2019variational} is a barycenter  yielded by optimizing the weighted sum of the forward KL divergence:
\begin{align*}
    \Tilde{q}_{\text{MoE}}(\bm{z}) &= \argmin\limits_{q(\bm{z})}  \sum_{m=1}^M \lambda_m D^{\text{reverse}}_{\text{KL}}( q_{\phi_m}||q) \\
    &= \argmin\limits_{q(\bm{z})}  \sum_{m=1}^M \lambda_m \Big[\underbrace{-\int  q_{\phi_m}(\bm{z}|\bm{x}_m) \log q(\bm{z}) d\bm{z}}_{\text{cross-entropy:} H(q_{\phi_m}, q)}\\
    & \quad \quad \quad \quad \quad + \underbrace{\int q_{\phi_m}(\bm{z}|\bm{x}_m)\log q_{\phi_m}(\bm{z}|\bm{x}_m) d\bm{z}}_{\text{negative entropy:}-H(q_{\phi_m})} \Big] \\
    &= \argmin\limits_{q(\bm{z})} -\int  \sum_{m=1}^M  \lambda_m  q_{\phi_m}(\bm{z}|\bm{x}_m) \log q(\bm{z}) d\bm{z} \\
    & \quad \quad \quad \quad \quad \quad \quad \quad \quad - \sum_{m=1}^M \lambda_m H(q_{\phi_m}(\bm{z}|\bm{x}_m)) \\
    &= \argmin\limits_{q(\bm{z})} H \left( \sum_{m=1}^M  \lambda_m  q_{\phi_m}(\bm{z}|\bm{x}_m), q(\bm{z})\right)
\end{align*}
The global optimum of minimizing the cross entropy between $\sum_{m=1}^M  \lambda_m  q_{\phi_m}(\bm{z}|\bm{x}_m)$ and $q(\bm{z})$ in the last line is attained at
$\Tilde{q}_{\text{MoE}}(\bm{z} | \bm{X}_{1:M})= \sum_{m=1}^M  \lambda_m  q_{\phi_m}(\bm{z}|\bm{x}_m)$, as the cross entropy is convex on $q$. The MoE is a special case when $\lambda_m = 1 /M , \ \forall m$. Unlike the aggregated distribution by PoE, the aggregated distribution by MoE is a valid probability measure by nature.

Here, we conclude that PoE and MoE are two barycenters with reverse and forward KL divergence as a divergence measure, respectively. However, due to the fact that KL divergence does not define a probability measure space, as it is unbounded and asymmetric, the resulting barycenter may not be a valid probability measure. 
\end{proof}

\subsection{A.3 \quad Proof of Remark 3}
\begin{proof}
We prove this by directly optimizing the weighted 2-Wasserstein distance, as it derives both $\Tilde{\bm{\mu}}$ and $\Tilde{\bm{\sigma}}$: 

\begin{align*}
     \Tilde{q}_{\mathcal{WB}}(\bm{z} | \bm{X}_{1:M}) &=\argmin\limits_{q} \sum_{m=1}^M \lambda_m \mathcal{W}_2^2(q_{\phi_m}, q)\\
     &= \argmin\limits_{\Tilde{\bm{\mu}}, \Tilde{\bm{\sigma}}} \sum_{m=1}^M \lambda_m \big[(\Tilde{\bm{\mu}} -  \bm{\mu}_m)^2 + \\
     & \quad \quad \quad \quad \quad \quad \quad \quad \quad(\tilde{\bm{\sigma}} - \bm{\sigma}_m)^2\big] 
\end{align*}
To improve readability, we define a function $\mathcal{L}(\Tilde{\bm{\mu}}, \Tilde{\bm{\sigma}})=\sum_{m=1}^M \lambda_m [(\Tilde{\bm{\mu}} -  \bm{\mu}_m)^2 + (\tilde{\bm{\sigma}} - \bm{\sigma}_m)^2]$. We then take the derivative of $\mathcal{L}(\Tilde{\bm{\mu}}, \Tilde{\bm{\sigma}})$ w.r.t. $\Tilde{\bm{\mu}}$ and $\Tilde{\bm{\sigma}}$, and then set them to zero: 
\begin{align*}
    \frac{\partial \mathcal{L}(\Tilde{\bm{\mu}}, \Tilde{\bm{\sigma}})}{\partial \Tilde{\bm{\mu}}} &= 2\sum_{m=1}^M \lambda_m  (\Tilde{\bm{\mu}} - \bm{\mu}_m) = 0 \\
     \frac{\partial \mathcal{L}(\Tilde{\bm{\mu}}, \Tilde{\bm{\sigma}})}{\partial \Tilde{\bm{\sigma}}} &= 2\sum_{m=1}^M \lambda_m  (\Tilde{\bm{\sigma}} - \bm{\sigma}_m) = 0 \\
     \Rightarrow &\begin{cases}
         \Tilde{\bm{\mu}} = \sum_{m=1}^M \lambda_m \bm{\mu}_m \\
         \Tilde{\bm{\sigma}} =\sum_{m=1}^M \lambda_m \bm{\sigma}_m. \\
     \end{cases}
\end{align*}

Alternatively, the same results can be derived by solving Eq.~(7) dimension by dimension for isotropic Gaussian with a diagonal covariance, where the solution is obvious. It is worth noting that the same results can also be derived by leveraging Proposition~\ref{lemma:1}, which turns out to be optimizing the squared 2-Wasserstein distance between the sought-after distribution and the mixture of unimodal (Gaussian) inference distributions. 


\end{proof}

\subsection{A.4 \quad MoPoE as a Barycenter}
Following the convention in Remark~\ref{remark:bures_wass}, MoPoE can be defined as a barycenter as follows:
\begin{equation}\nonumber
    \begin{split}
         &\quad \Tilde{q}_{\mathcal{MWB}}(\bm{z} | \bm{X}) = \argmin_{q} \sum_{\bm{X}_k 
     \in \mathcal{P}_M(\bm{X})}\lambda_k  D_{\text{KL}}(\Tilde{q}_{\mathcal{WB}} || q) \\
     &\text{subject to} \quad \Tilde{q}_{\mathcal{WB}}(\bm{z} | \bm{X}_{k})  =\argmin\limits_{q} \sum_{\bm{x}_j \in \bm{X}_k} \lambda_j D_{\text{KL}}(q_{\phi_j}, q).
    \end{split}
    \end{equation} 
Leveraging Proposition~\ref{lemma:1}, it is trivial to show MoPoE can be simplified to the form defined in~\citet{sutter2021generalized}:
\begin{align*}
    \Tilde{q}_{\text{MoPoE}} = \argmin_q D_{\text{KL}}\left(\frac{1}{2^M} \sum_{\bm{X}_k \in \mathcal{P}_M(\bm{X})}  \prod_{\bm{x}_j \in \bm{X}_k }q_{\phi_j} , q \right),
\end{align*}
where the optimum attains when $\Tilde{q}_{\text{MoPoE}}(\bm{z}|\bm{X}_{1:M}) = \frac{1}{2^M} \sum_{\bm{X}_k \in \mathcal{P}_M(\bm{X})}  \prod_{\bm{x}_j \in \bm{X}_k }q_{\phi_j}(\bm{z}|\bm{x}_j)$. 

\section{B \quad Additional Experimental Results}
Here, we provide additional experimental details (e.g., hyperparameters) as well as additional quantitative and qualitative results for different datasets.
For all the experiments, we used the same neural network architectures as outlined in~\citet{sutter2021generalized} for a fair comparison. Unless otherwise specified, the experiments were repeated five times, with the means and standard deviations reported. Following the protocols outlined in~\citet{sutter2021generalized}, the three evaluation metrics (i.e., the quality of the learned latent representations, the coherence of the generated samples and the log-likelihood on the test set) were computed as follows. First, the quality of the learned latent representations was evaluated using a logistic regression classifier that was trained on 500 samples from the training set. The reported results are the average performances of the trained classifier on the test set by taking the learned latent representations as inputs. The coherence of the generated samples was evaluated by classifying if the generated samples were from certain modalities. For this purpose, we pretrained a classifier (which has the same architectures as the unimodal encoders) for every modality to classify if a generated sample is coherent. Let us take the condition generation of MNIST digits when taking the text as inputs on the MINIST-SVHN-TEXT dataset as an example.  The coherence of the generated MNIST digits is calculated as the ratio of coherent samples classified as MNIST by the pretrained classifier divided by the total number of generated samples. Third, the average log-likelihood on the test set is calculated by averaging the log-likelihoods of multiple generated samples for each input.

\subsection{B.1 \quad PolyMNIST}
\subsubsection{Dataset details.}
The PolyMNIST contains five different modalities by mixing the MNIST digit with a random crop of size $28 \times 28$ from five different large background images\footnote{Urls for five background images: \\
\url{https://people.sc.fsu.edu/~jburkardt/data/jpg/fractal_tree.jpg}, \\ \url{https://upload.wikimedia.org/wikipedia/commons/f/f4/The_Scream.jpg}, \\ \url{http://links.uwaterloo.ca/Repository/TIF/lena3.tif}, \\ \url{https://people.sc.fsu.edu/~jburkardt/data/jpg/star_field.jpg},  \\ \url{https://people.sc.fsu.edu/~jburkardt/data/jpg/shingles.jpg}}. 

\subsubsection{Experiment setup.}
We trained all models for 300 epochs using an Adam optimization~\cite{kingma2014adam} with an initial learning rate of 0.001. The weight balance parameter of the KL divergence was set to 2.5. The batch size was set to 256. The neural network architectures were the same as those used in~\citet{sutter2021generalized} with a latent dim of 512 for all modalities. 

\subsubsection{Additional qualitative results.}
\begin{figure*}[ht]
    \centering
    \begin{subfigure}[t]{0.24\textwidth}
        \centering
        \includegraphics[width=0.95\textwidth]{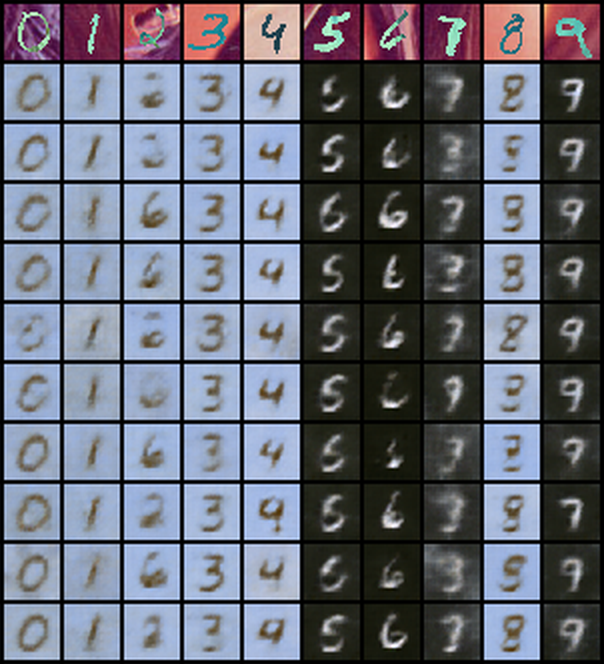}
        \caption{$\mathcal{MWB}$-VAE}
    \end{subfigure}%
    \begin{subfigure}[t]{0.24\textwidth}
        \centering
        \includegraphics[width=0.95\textwidth]{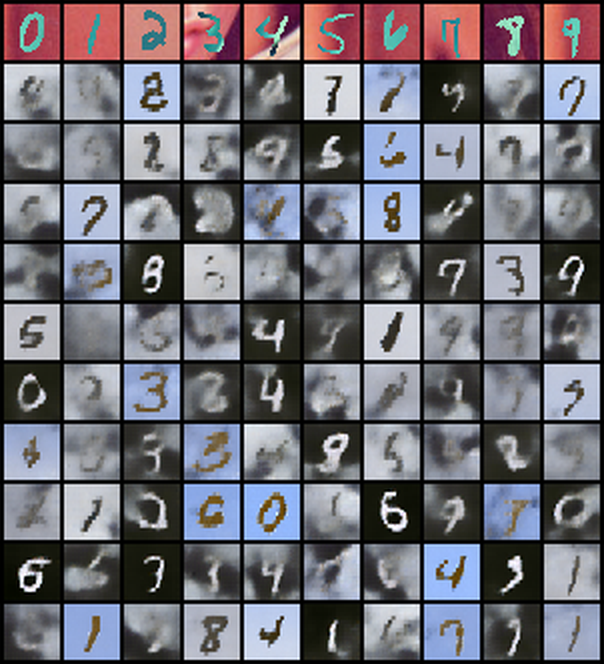}
        \caption{PoE-VAE}
    \end{subfigure}
     \begin{subfigure}[t]{0.24\textwidth}
        \centering
        \includegraphics[width=0.95\textwidth]{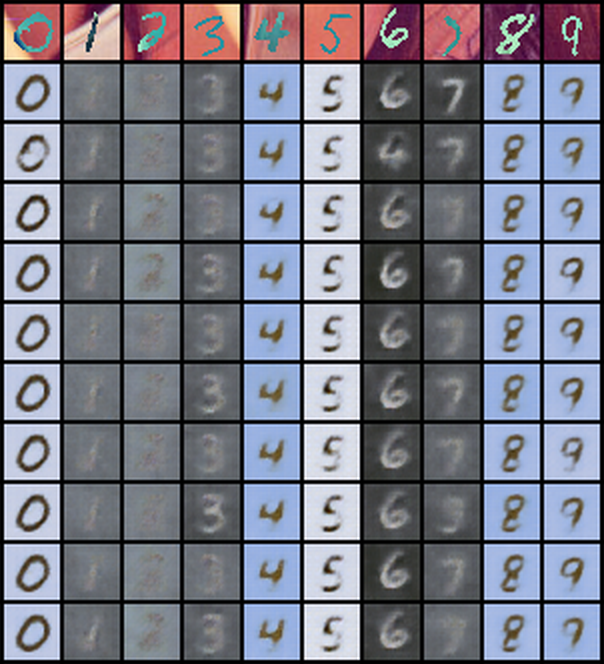}
        \caption{MoE-VAE}
    \end{subfigure}
     \begin{subfigure}[t]{0.24\textwidth}
        \centering
        \includegraphics[width=0.95\textwidth]{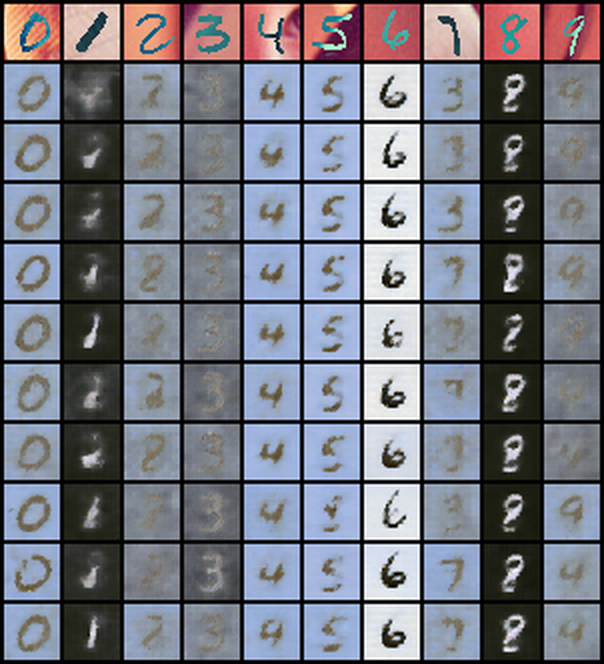}
        \caption{MoPoE-VAE}
    \end{subfigure}
    \caption{Conditionally generated samples of the first modality (from the second to the last rows) given the respective test example from the third modality (first row). For each column, we draw distinct samples from the approximate joint posterior, which should generate the same digits but be expected to show stylistic variations.}
    \label{fig:polymnist:m2_m0}
\end{figure*}

\begin{figure*}[ht]
    \centering
    \begin{subfigure}[t]{0.24\textwidth}
        \centering
        \includegraphics[width=0.95\textwidth]{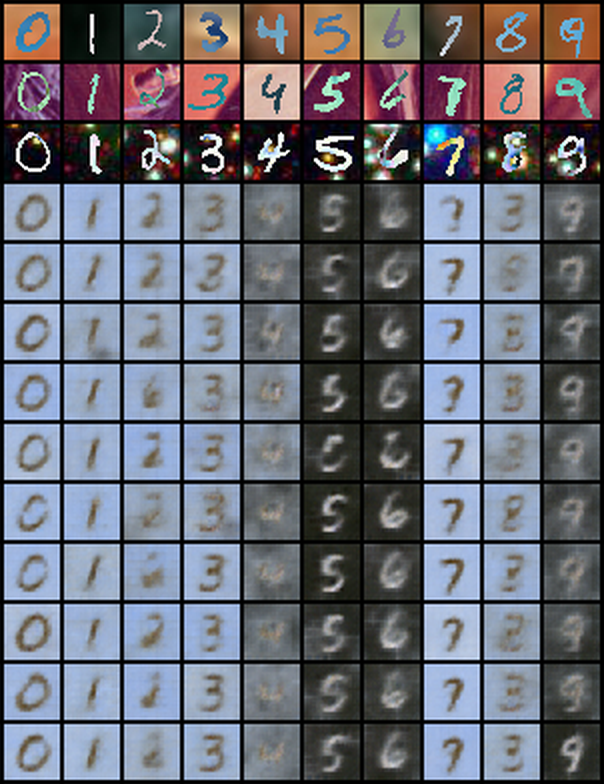}
        \caption{$\mathcal{MWB}$-VAE}
    \end{subfigure}%
    \begin{subfigure}[t]{0.24\textwidth}
        \centering
        \includegraphics[width=0.95\textwidth]{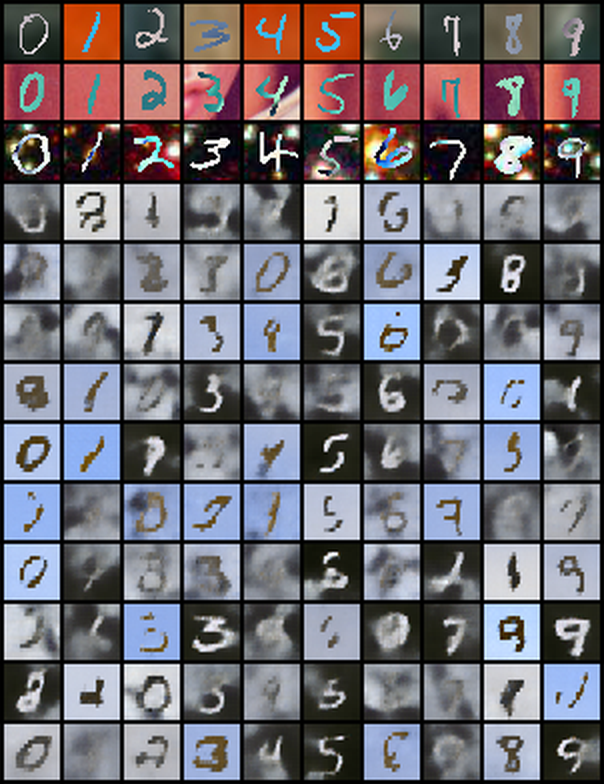}
        \caption{PoE-VAE}
    \end{subfigure}
     \begin{subfigure}[t]{0.24\textwidth}
        \centering
        \includegraphics[width=0.95\textwidth]{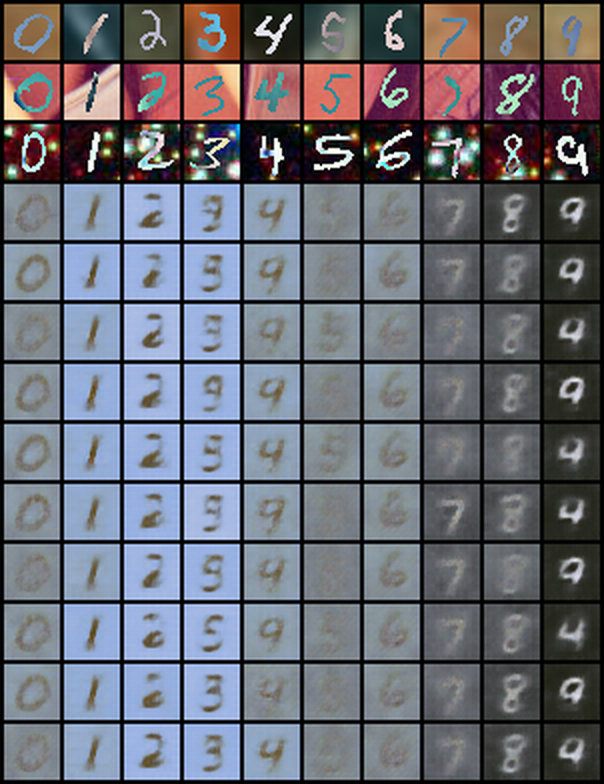}
        \caption{MoE-VAE}
    \end{subfigure}
     \begin{subfigure}[t]{0.24\textwidth}
        \centering
        \includegraphics[width=0.95\textwidth]{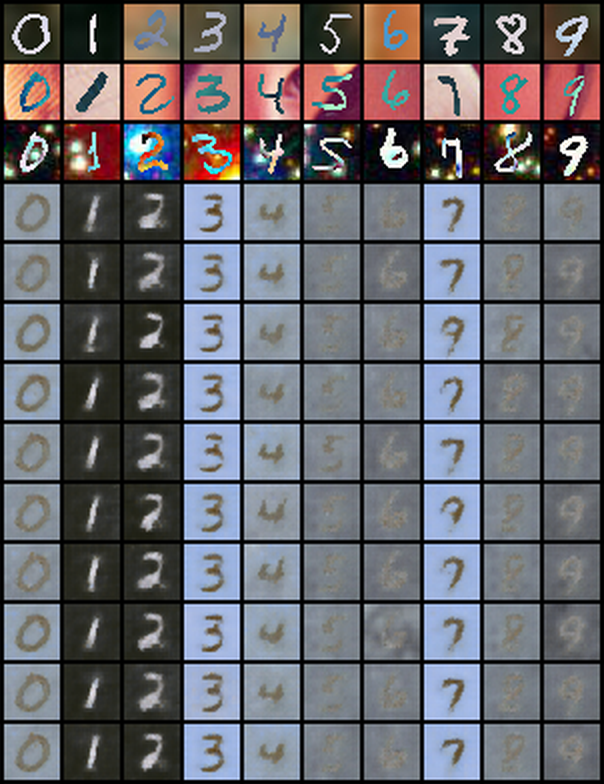}
        \caption{MoPoE-VAE}
    \end{subfigure}
    \caption{Conditionally generated samples of the first modality (from the fourth to the last rows) given the respective test example from the second, third, and fourth modalities (first three rows). For each column, we draw distinct samples from the approximate joint posterior, which should generate the same digits but be expected to show stylistic variations.}
    \label{fig:polymnist:m_1_m2_m3_m0}
\end{figure*}

\begin{figure*}[ht]
    \centering
    \begin{subfigure}[t]{0.24\textwidth}
        \centering
        \includegraphics[width=0.95\textwidth]{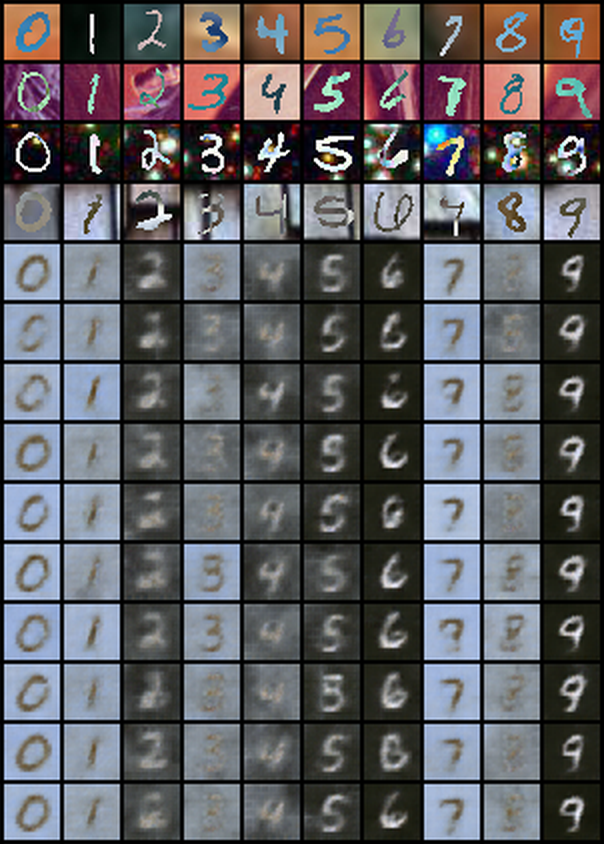}
        \caption{$\mathcal{MWB}$-VAE}
    \end{subfigure}%
    \begin{subfigure}[t]{0.24\textwidth}
        \centering
        \includegraphics[width=0.95\textwidth]{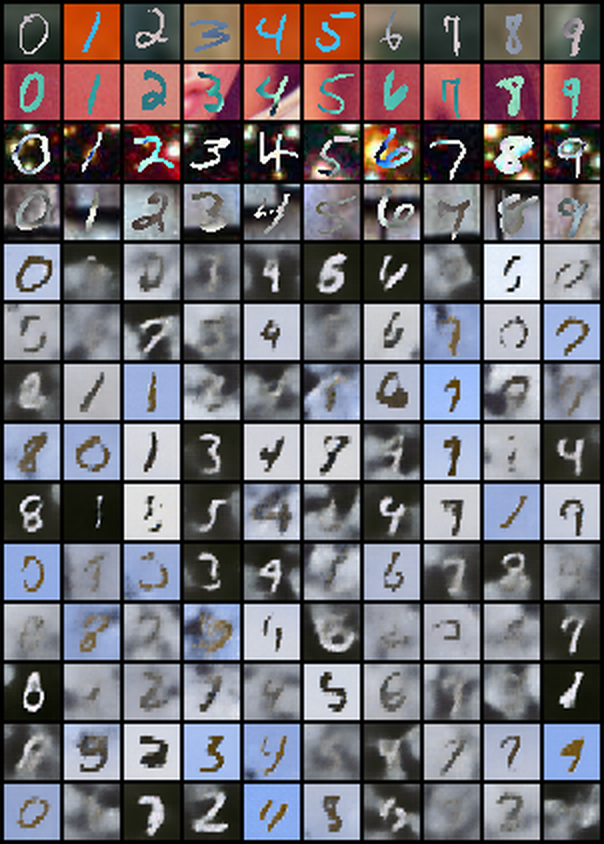}
        \caption{PoE-VAE}
    \end{subfigure}
     \begin{subfigure}[t]{0.24\textwidth}
        \centering
        \includegraphics[width=0.95\textwidth]{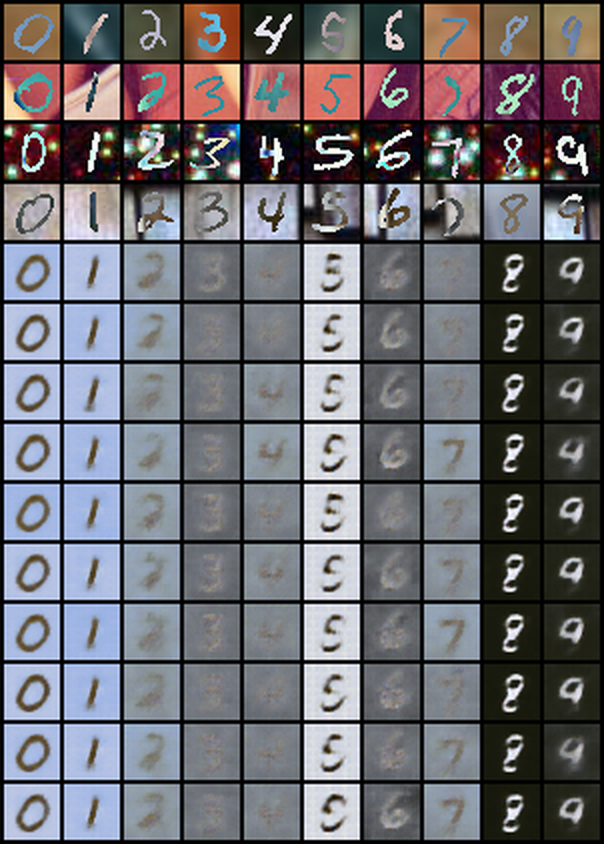}
        \caption{MoE-VAE}
    \end{subfigure}
     \begin{subfigure}[t]{0.24\textwidth}
        \centering
        \includegraphics[width=0.95\textwidth]{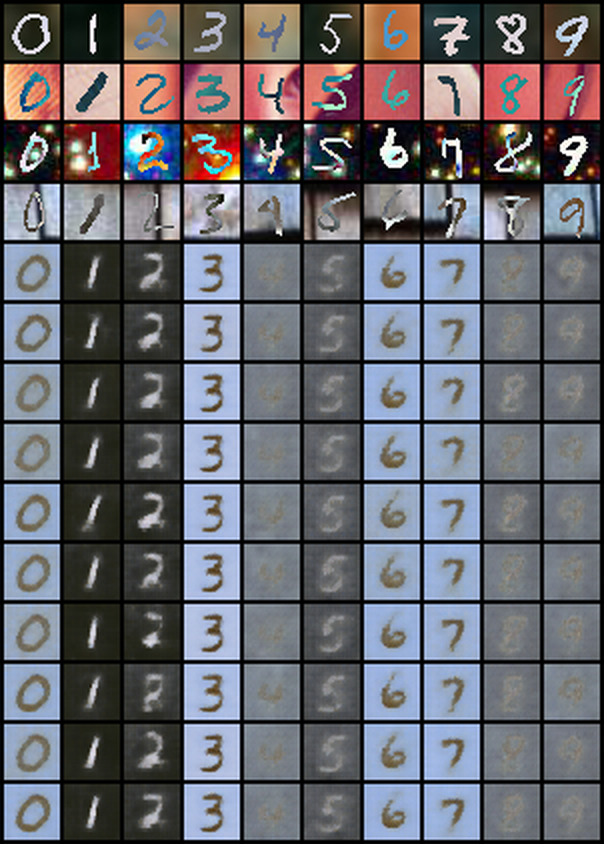}
        \caption{MoPoE-VAE}
    \end{subfigure}
    \caption{Conditionally generated samples of the first modality (from the fifth to the last rows) given the respective test example from the rest four modalities (first four rows). For each column, we draw distinct samples from the approximate joint posterior, which should generate the same digits but be expected to show stylistic variations.}
    \label{fig:polymnist:m_1_m2_m3_m4_m0}
\end{figure*}

We show additional qualitative results of the proposed $\mathcal{MWB}$-VAE in comparison to PoE-VAE, MoE-VAE, and MoPoE by giving different 
modalities as input (see Figs.~\ref{fig:polymnist:m2_m0},~\ref{fig:polymnist:m_1_m2_m3_m0}, and~\ref{fig:polymnist:m_1_m2_m3_m4_m0}).  

\subsection{B.2 \quad MNIST-SVHN-TEXT}
\subsubsection{Dataset details.}
MNIST digit, text, and SVHN~\cite{netzer2011reading}. The MNIST digit and text are two clean modalities, whereas SVHN is comprised of noisy images. Folliwing~\citet{sutter2021generalized}, 20 triples were generated per set using a many-to-many mapping.

\subsubsection{Experiment setup.}
We trained all models for 150 epochs using an Adam optimization~\cite{kingma2014adam} with an initial learning rate of 0.001. The weight balance parameter of the KL divergence was set to 5.0. The batch size was set to 256. The neural network architectures were the same as used in~\citet{sutter2021generalized} with a latent dim of 20 for all modalities. 

\subsubsection{Additional qualitative results.}
\begin{figure*}[ht]
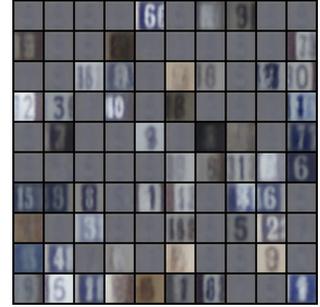
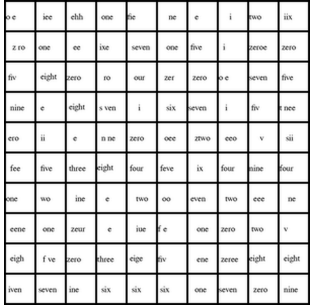

    \centering
    \begin{subfigure}[t]{0.24\textwidth}
        \centering
        \includegraphics[width=0.95\textwidth]{figures/MNISTSVHNTEXT/unconditional/MWB/MNIST.pdf}
        \caption{$\mathcal{MWB}$-VAE: MNIST}
    \end{subfigure}%
    \begin{subfigure}[t]{0.24\textwidth}
        \centering
        \includegraphics[width=0.95\textwidth]{figures/MNISTSVHNTEXT/unconditional/PoE/MNIST.pdf}
        \caption{PoE-VAE: MNIST}
    \end{subfigure}
     \begin{subfigure}[t]{0.24\textwidth}
        \centering
        \includegraphics[width=0.95\textwidth]{figures/MNISTSVHNTEXT/unconditional/MoE/MNIST.pdf}
        \caption{MoE-VAE: MNIST}
    \end{subfigure}
     \begin{subfigure}[t]{0.24\textwidth}
        \centering
        \includegraphics[width=0.95\textwidth]{figures/MNISTSVHNTEXT/unconditional/MoPoE/MNIST.pdf}
        \caption{MoPoE-VAE: MNIST}
    \end{subfigure}

      \begin{subfigure}[t]{0.24\textwidth}
        \centering
        \includegraphics[width=0.95\textwidth]{figures/MNISTSVHNTEXT/unconditional/MWB/SVHN.pdf}
        \caption{$\mathcal{MWB}$-VAE: SVHN}
    \end{subfigure}%
    \begin{subfigure}[t]{0.24\textwidth}
        \centering
        \includegraphics[width=0.95\textwidth]{figures/MNISTSVHNTEXT/unconditional/PoE/SVHN.pdf}
        \caption{PoE-VAE: SVHN}
    \end{subfigure}
     \begin{subfigure}[t]{0.24\textwidth}
        \centering
        \includegraphics[width=0.95\textwidth]{figures/MNISTSVHNTEXT/unconditional/MoE/SVHN.pdf}
        \caption{MoE-VAE: SVHN}
    \end{subfigure}
     \begin{subfigure}[t]{0.24\textwidth}
        \centering
        \includegraphics[width=0.95\textwidth]{figures/MNISTSVHNTEXT/unconditional/MoPoE/SVHN.pdf}
        \caption{MoPoE-VAE: SVHN}
    \end{subfigure}

    \begin{subfigure}[t]{0.24\textwidth}
        \centering
        \includegraphics[width=0.95\textwidth]{figures/MNISTSVHNTEXT/unconditional/MWB/TEXT.pdf}
        \caption{$\mathcal{MWB}$-VAE: Text}
    \end{subfigure}%
    \begin{subfigure}[t]{0.24\textwidth}
        \centering
        \includegraphics[width=0.95\textwidth]{figures/MNISTSVHNTEXT/unconditional/PoE/TEXT.pdf}
        \caption{PoE-VAE: Text}
    \end{subfigure}
     \begin{subfigure}[t]{0.24\textwidth}
        \centering
        \includegraphics[width=0.95\textwidth]{figures/MNISTSVHNTEXT/unconditional/MoE/TEXT.pdf}
        \caption{MoE-VAE: Text}
    \end{subfigure}
     \begin{subfigure}[t]{0.24\textwidth}
        \centering
        \includegraphics[width=0.95\textwidth]{figures/MNISTSVHNTEXT/unconditional/MoPoE/TEXT.pdf}
        \caption{MoPoE-VAE: Text}
    \end{subfigure}
    
    \caption{Qualitative comparison of randomly generated MNIST-SVHN-Text samples: (a) - (d) MNIST digit, (e) - (h) SVHN, and (i) - (l) Text.}
    \label{fig:mnist_text:random}
\end{figure*}

\begin{figure*}[ht]
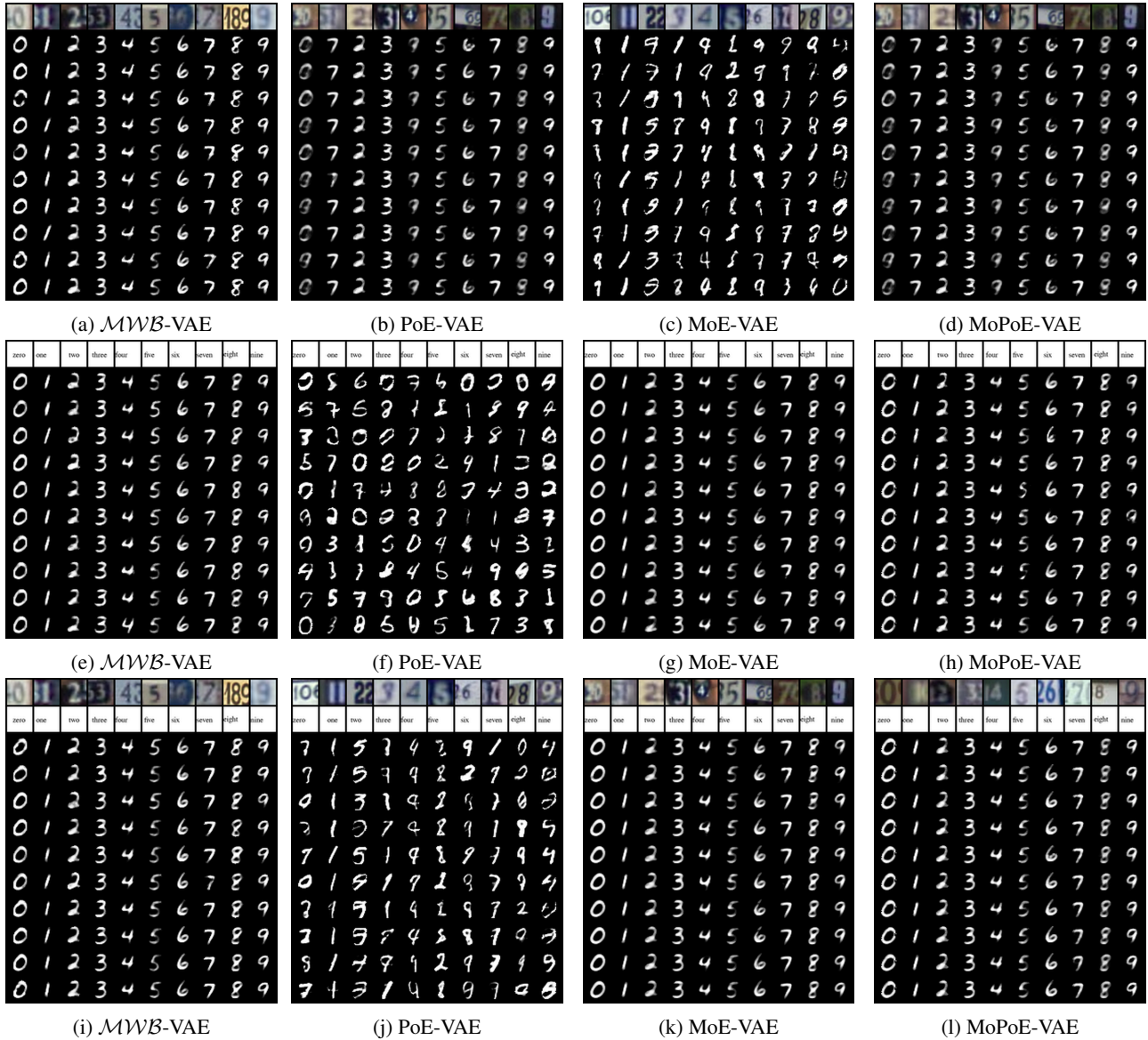

    \centering
    \begin{subfigure}[t]{0.24\textwidth}
        \centering
        \includegraphics[width=0.95\textwidth]{figures/MNISTSVHNTEXT/conditional/S_to_M/MWB.pdf}
        \caption{$\mathcal{MWB}$-VAE}
    \end{subfigure}%
    \begin{subfigure}[t]{0.24\textwidth}
        \centering
        \includegraphics[width=0.95\textwidth]{figures/MNISTSVHNTEXT/conditional/S_to_M/MoE.pdf}
        \caption{PoE-VAE}
    \end{subfigure}
     \begin{subfigure}[t]{0.24\textwidth}
        \centering
        \includegraphics[width=0.95\textwidth]{figures/MNISTSVHNTEXT/conditional/S_to_M/PoE.pdf}
        \caption{MoE-VAE}
    \end{subfigure}
     \begin{subfigure}[t]{0.24\textwidth}
        \centering
        \includegraphics[width=0.95\textwidth]{figures/MNISTSVHNTEXT/conditional/S_to_M/MoE.pdf}
        \caption{MoPoE-VAE}
    \end{subfigure}

      \begin{subfigure}[t]{0.24\textwidth}
        \centering
        \includegraphics[width=0.95\textwidth]{figures/MNISTSVHNTEXT/conditional/T_to_M/MWB.pdf}
        \caption{$\mathcal{MWB}$-VAE}
    \end{subfigure}%
    \begin{subfigure}[t]{0.24\textwidth}
        \centering
        \includegraphics[width=0.95\textwidth]{figures/MNISTSVHNTEXT/conditional/T_to_M/PoE.pdf}
        \caption{PoE-VAE}
    \end{subfigure}
     \begin{subfigure}[t]{0.24\textwidth}
        \centering
        \includegraphics[width=0.95\textwidth]{figures/MNISTSVHNTEXT/conditional/T_to_M/MoE.pdf}
        \caption{MoE-VAE}
    \end{subfigure}
     \begin{subfigure}[t]{0.24\textwidth}
        \centering
        \includegraphics[width=0.95\textwidth]{figures/MNISTSVHNTEXT/conditional/T_to_M/MoPoE.pdf}
        \caption{MoPoE-VAE}
    \end{subfigure}

    \begin{subfigure}[t]{0.24\textwidth}
        \centering
        \includegraphics[width=0.95\textwidth]{figures/MNISTSVHNTEXT/conditional/ST_to_M/MWB.pdf}
        \caption{$\mathcal{MWB}$-VAE}
    \end{subfigure}%
    \begin{subfigure}[t]{0.24\textwidth}
        \centering
        \includegraphics[width=0.95\textwidth]{figures/MNISTSVHNTEXT/conditional/ST_to_M/PoE.pdf}
        \caption{PoE-VAE}
    \end{subfigure}
     \begin{subfigure}[t]{0.24\textwidth}
        \centering
        \includegraphics[width=0.95\textwidth]{figures/MNISTSVHNTEXT/conditional/ST_to_M/MoE.pdf}
        \caption{MoE-VAE}
    \end{subfigure}
     \begin{subfigure}[t]{0.24\textwidth}
        \centering
        \includegraphics[width=0.95\textwidth]{figures/MNISTSVHNTEXT/conditional/ST_to_M/MoPoE.pdf}
        \caption{MoPoE-VAE}
    \end{subfigure}
    
    \caption{Qualitative comparison of conditionally generated MNIST digits given (a) - (d)  SVHN, (e) - (h) Text, and (i) - (l) SVHN-Text. For each column, we draw distinct samples from the approximate joint posterior, which should generate the same digits.}
    \label{fig:mnist_text:conditional}
\end{figure*}

We show additional qualitative results of randomly and conditionally generated samples from the proposed $\mathcal{MWB}$-VAE in comparison to PoE-VAE, MoE-VAE, and MoPoE (see Figs.~\ref{fig:mnist_text:random} and~\ref{fig:mnist_text:conditional}).  

\subsection{B.3 \quad Bimodal CelebA}
\subsubsection{Dataset details.}
The bimodal CelebA dataset consists of human face images with 40 different text attributes associated with them. The text modality consists of attribute strings, which are present in a face image, separated by commas. The text modality is more challenging. This is because an attribute string is not present in the text modalities if the attribute is not present in a face image.  

\subsubsection{Experimental setup.}
We trained all models for 150 epochs using an Adam optimization~\cite{kingma2014adam} with an initial learning rate of 0.001. The weight balance parameter of the KL divergence was set to 2.5. The batch size was set to 256. The neural network architectures were the same as used in~\citet{sutter2021generalized} with a latent dim of 32 for all modalities. Similar to~\citet{sutter2021generalized}, an additional modality-specific latent space with the same dim pf 32 was added to each modality, resulting in a total latent dimension of 64 per modality. 

\subsubsection{Additional results.}
\begin{figure*}[ht]
    \centering
    \begin{subfigure}[t]{1.0\textwidth}
        \centering
        \includegraphics[width=0.8\textwidth]{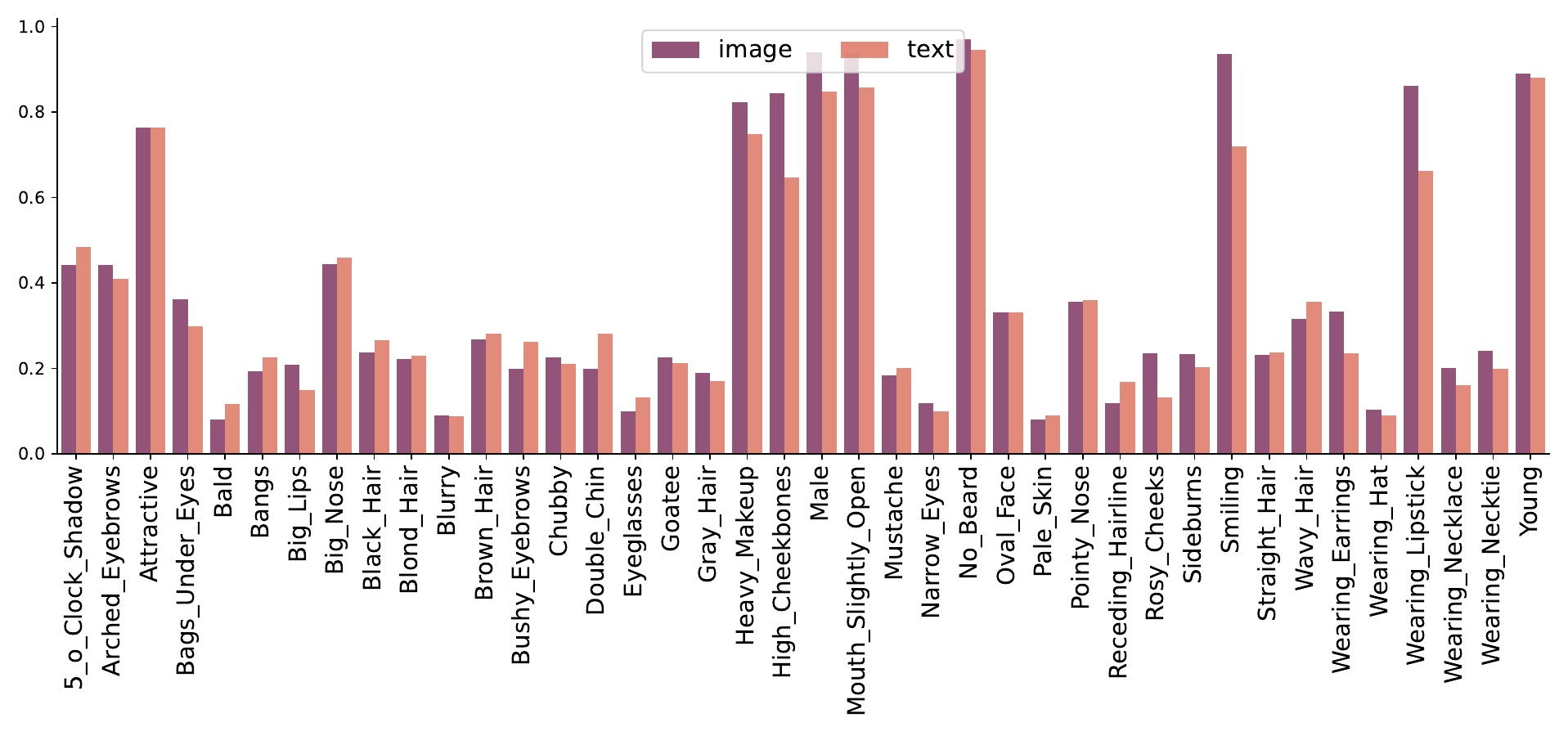}
        \caption{Input modality: image}
    \end{subfigure}%
    \\
    \begin{subfigure}[t]{1.0\textwidth}
        \centering
        \includegraphics[width=0.8\textwidth]{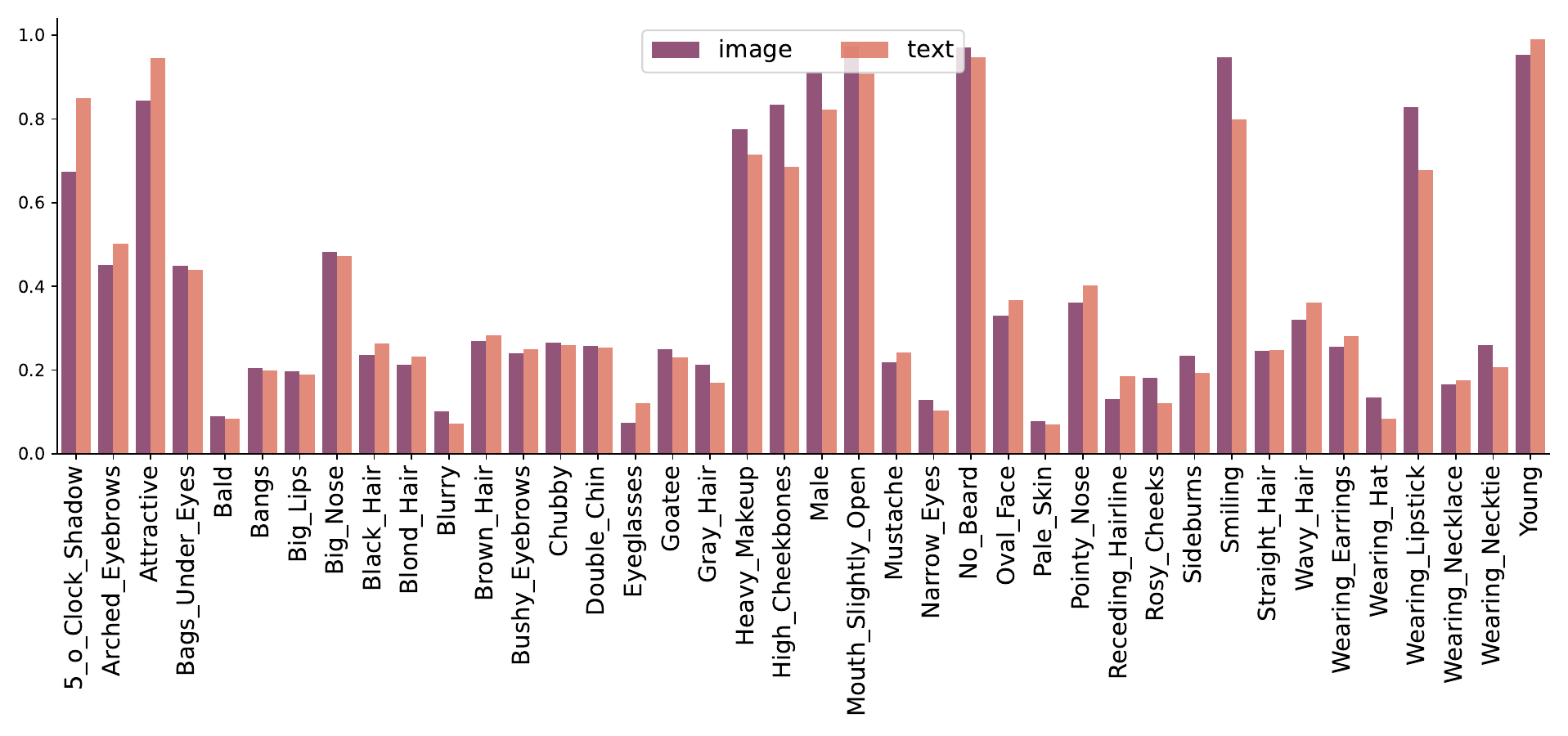}
        \caption{Input modality: text}
    \end{subfigure}
    \\
     \begin{subfigure}[t]{1.0\textwidth}
        \centering
        \includegraphics[width=0.8\textwidth]{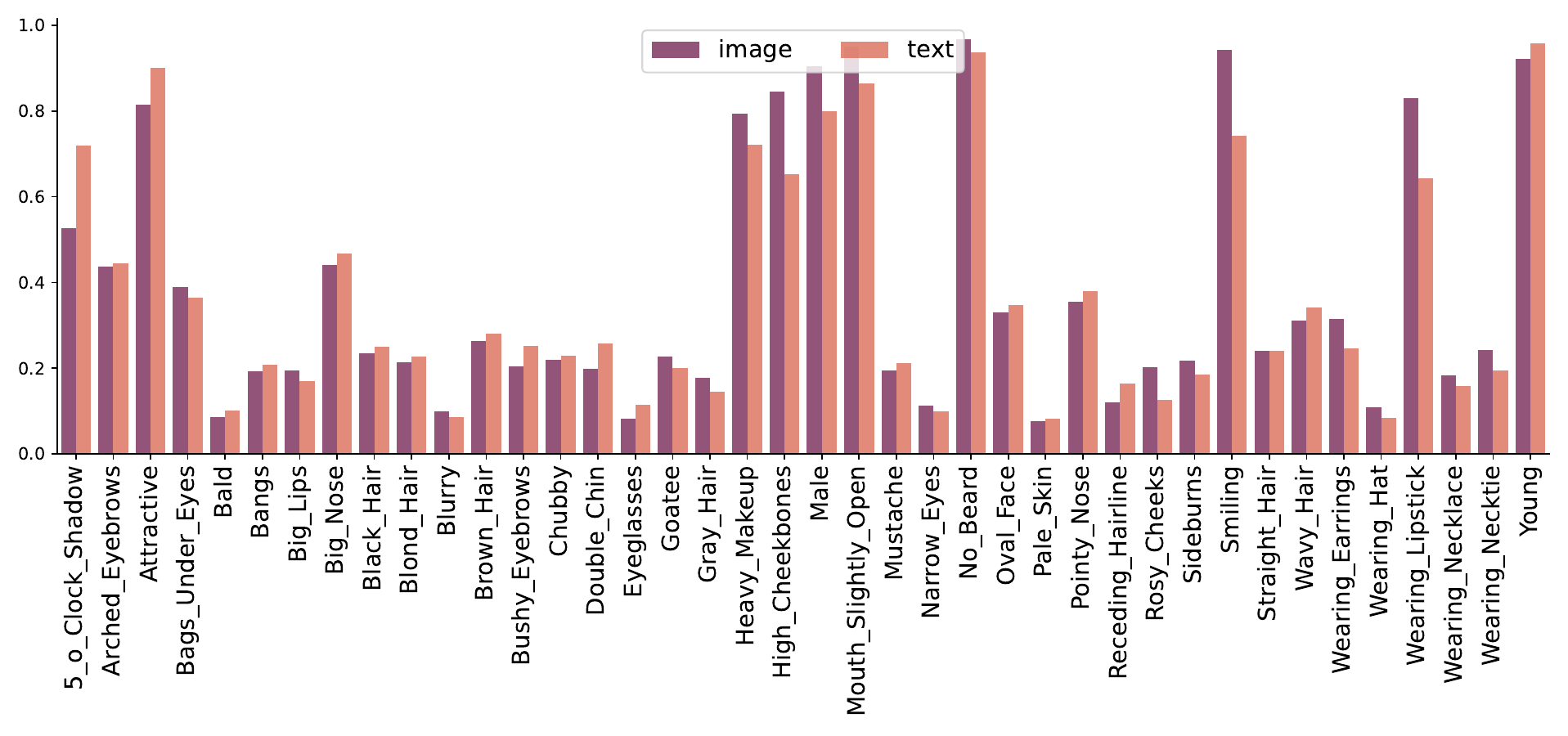}
        \caption{Input modality: image \& text}
    \end{subfigure}
    \caption{The coherence of the generated face images and text attributes on the bimodal CelebA dataset using $\mathcal{MWB}$-VAE by taking different modalities as inputs: (a) image, (b) text, and (c) image \& text. }
    \label{fig:celebaA:generation}
\end{figure*}

\begin{figure*}[ht]
    \centering
    \includegraphics[width=0.8\textwidth]{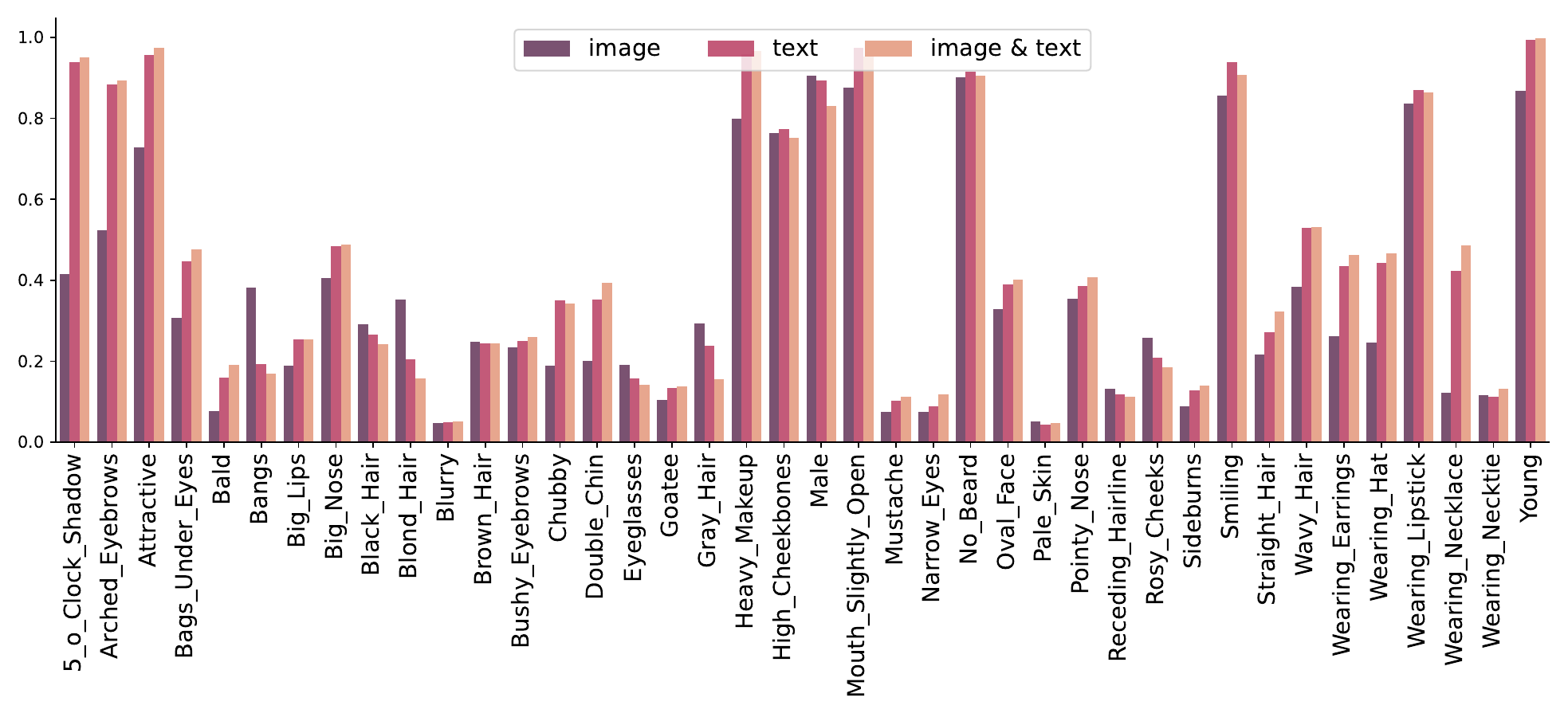}
    \caption{The quality of the learned latent representation for the bimodal CelebA dataset using $\mathcal{MWB}$-VAE, given image, text, or both as inputs.}
    \label{fig:celebaA:latent}
\end{figure*}

\begin{figure*}[ht]
    \centering
    \includegraphics[width=0.95\textwidth]{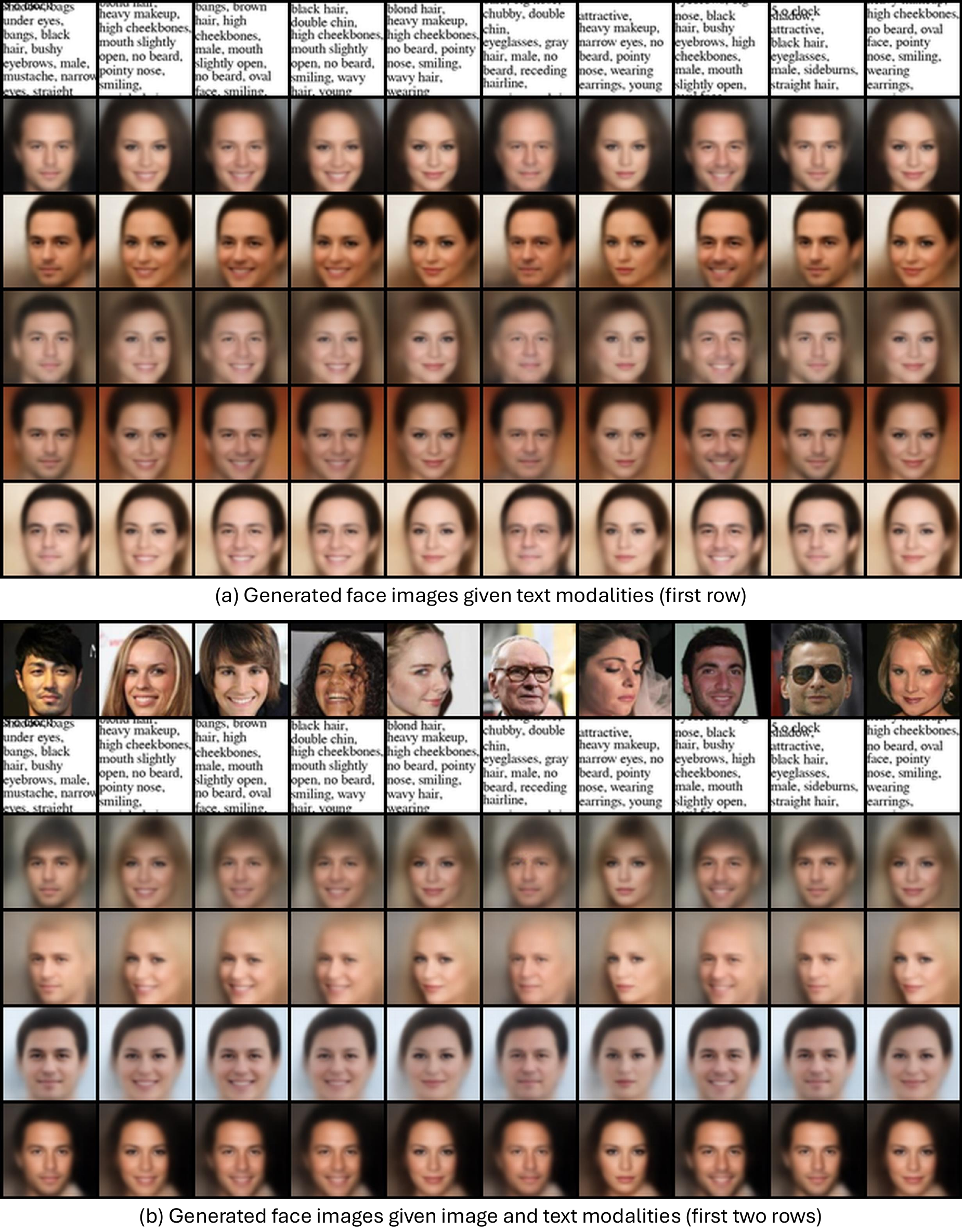}
    \caption{The conditionally generated human face images given (a) text and (b) text \& image as input modalities.}
    \label{fig:celebaA:vis}
\end{figure*}

We provide the distribution of the evaluations for each attribute in Fig.~\ref{fig:celebaA:generation} (generation coherence) and Fig.~\ref{fig:celebaA:latent} (latent representation quality). $\mathcal{MWB}$-VAE showed good performance for most of the large attributes, although there is room for improvement for smaller attributes that are inherently more challenging. Similar trends were also observed in the conditionally generated samples, as shown in Fig.~\ref{fig:celebaA:vis}.

\end{document}